\PassOptionsToPackage{hyphens}{url}
\documentclass[11pt]{article}

\usepackage[final]{acl}

\usepackage{times}
\usepackage{latexsym}
\usepackage[T1]{fontenc}
\usepackage[utf8]{inputenc}
\usepackage{microtype}
\usepackage{inconsolata}

\usepackage{amsmath}
\usepackage{amssymb}
\usepackage{mathtools}
\usepackage{amsthm}

\usepackage{graphicx}
\usepackage{subcaption}
\usepackage{booktabs}
\usepackage{multirow}
\usepackage{arydshln}
\usepackage{float}

\usepackage{xspace}
\usepackage[capitalize,noabbrev]{cleveref}
\usepackage[textsize=tiny]{todonotes}

\theoremstyle{plain}

\theoremstyle{definition}

\theoremstyle{remark}

\newcommand{\ours}{\textsc{InsertQuant}\xspace}

\title{Massive Spikes in LLMs are Bias Vectors: Mechanistic Uncovering and Spike-Free Quantization}

\author{Yung-Chin Chen\textsuperscript{1} \quad 
  Chung Peng Lee\textsuperscript{1} \quad 
  Ze-Wei Liou\textsuperscript{1} \quad 
  Naveen Verma\textsuperscript{1 2} \\
  \textsuperscript{1}Princeton University, NJ, USA \quad \textsuperscript{2}EnCharge AI, CA, USA \\
  \texttt{\{yc9182, cl6486, zl3193, nverma\}@princeton.edu} \\}

\begin{document}
\maketitle

\begin{abstract}
  Massive activation spikes in Large Language Models (LLMs) severely degrade quantization by increasing dynamic range requirements. While prior hypotheses characterize these as high-level \textbf{scalar biases}, we argue that they are merely the scalar intermediates of rigid, structural \textbf{vector biases} in the spike-carrying tokens. We show that these tokens converge to constant vectors \textit{after normalization} that drive the attention sink and value-state drain mechanisms. We geometrically substantiate this by analyzing the coordination of projection weights: $W_K$ contrastively amplifies the vector, $W_Q$ aligns semantic tokens toward it, and $W_V$ projects it into the spectral null-space. Furthermore, we reveal that the model actively preserves these structural biases against Rotary Positional Embedding (RoPE) perturbations by localizing them in ``zones of rotational stability'' utilizing low-frequency bands and coherent channel pairs. Leveraging this, we propose \ours, a post-training quantization (PTQ) framework that clamps spikes and restores their function via pre-computed \textbf{template vectors}. This renders activations strictly spike-free, enabling robust low-bit quantization with high fidelity.
  \ours achieves parity with state-of-the-art per-tensor quantization methods on LLMs and uniquely generalizes beyond text to other modalities such as ViTs.
\end{abstract}

\section{Introduction}

\begin{figure}[t]
    \centering
    \includegraphics[width=0.95\linewidth]{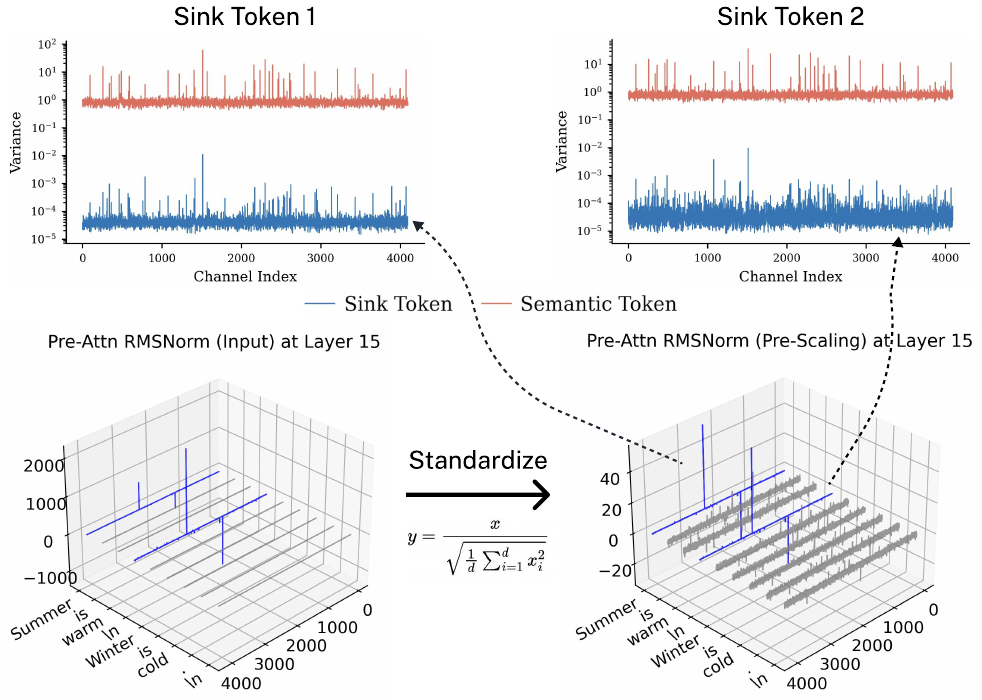}
    \caption{\textbf{Visualizing the Bias Vector Hypothesis.} \textbf{Bottom:} RMSNorm standardizes extreme input spikes (e.g., ``\texttt{\textbackslash n}'') into a stable direction. \textbf{Top:} While semantic tokens vary significantly to encode information (high variance), spike-carrying tokens converge to rigid \textbf{Bias Vector} ($\mathbf{b}$) after RMSNorm, exhibiting negligible variance independent of the input sequence.}
    \label{fig:channelwise-variance}
\end{figure}

The rapid progress of Large Language Models (LLMs) has unlocked remarkable capabilities but imposed severe computational and memory demands. The increasing scale of LLMs requires compatibility with efficient hardware, critically enabled by quantization. However, the efficacy of activation quantization is severely hindered by extreme activation values. While existing literature has effectively mitigated \textbf{Channel-wise Outliers} \citep{LLMINT8} (channels consistently large across all tokens) via smoothing \citep{Smoothquant} or rotation \citep{Quarot}, a distinct and more disruptive phenomenon remains: \textbf{Massive Spikes}.
These spikes manifest as extreme values localized to both specific channels and specific token semantics, routinely exhibiting magnitudes thousands of times larger than typical values
. They severely stretch the activation dynamic range and exacerbate quantization error, especially for hardware-efficient static per-tensor schemes.


Recent literature has moved beyond treating these spikes as numerical aberrations, proposing functional interpretations such as the \textit{Bias Hypothesis} \citep{MassiveActivations, SystematicOutliers}. However, these works interpret spikes incompletely as high-level \textit{scalar biases}. In this work, we advance this understanding from scalar biases to complete \textbf{vector biases} corresponding to the entire token. This \textit{Bias Vector Hypothesis} reveals that the bias exhibits far stronger rigidity and deterministic structure than previously observed, yielding critical benefits for (1) deeper understanding of model behaviors and (2) enhanced activation quantization. Regarding \textbf{deeper understanding}, the rigid vector structure more clearly exposes the mechanistic purpose of the bias and its precise function in creating \textit{Attention Sinks} and \textit{Value-state Drains}. Regarding \textbf{quantization}, this insight allow us to extend beyond previous approaches limited to text modalities, establishing a more generalized framework.


We empirically analyze the processing of this highly structured bias vector within the self-attention mechanism. We quantify how the vector drives the attention sink phenomenon through the coordination of projection weights: the Key ($W_K$) and Query ($W_Q$) projections interact to isolate the vector as a global attractor, while the Value ($W_V$) projection projects the vector to the null-space to mask the sink token's contribution to the residual stream. Furthermore, we show that this rigid vector structure is actively preserved against perturbations by localizing in specific stability zones defined by the channel-dependent properties of Rotary Positional Embeddings (RoPE) \cite{Roformer}.

Leveraging this \textit{Bias Vector Hypothesis}, we propose \ours, a post-training quantization (PTQ) framework that decouples these structural biases from content tokens. Since the spike is simply the precursor to a static bias vector, it does not require runtime computation. Instead, we surgically clamp spike-carrying tokens to zero prior to matrix multiplication---rendering the activation space strictly spike-free---and functionally restore them by inserting pre-computed \textbf{template vectors} in subsequent layers. This approach aligns the quantization strategy with the model's intrinsic mechanics, enabling high-performance low-bit inference. Crucially, because \ours operates on the embedding space rather than relying on heuristic token identities, it generalizes effectively across modalities like Vision Transformers (ViTs).

Our contributions are as follows:
\begin{itemize}
    \item \textbf{Bias Vector Hypothesis:} We establish that massive spikes are the scalar intermediates of rigid bias vectors. We show that while spike magnitudes fluctuate, their normalized direction remains invariant, serving as a structural model component.
    \item \textbf{Mechanistic Analysis:} We provide a quantitative geometric analysis of how bias vectors drive the attention sink and value-state drain mechanism. We detail their interaction with projection weights ($W_K, W_Q, W_V$) and identify their localization in rotational stability zones---specifically utilizing low-frequency bands and coherent channel pairs---as strong corroboration that the model actively preserves these rigid structures.
    \item \textbf{\ours Framework:} We propose the first PTQ framework to exploit the systematic nature of spikes, establishing a unified cross-modality solution. By treating these artifacts as pre-computable bias vectors, \ours surgically eliminates them via clamping and restoration, achieving high-fidelity low-bit quantization accuracy across both LLMs and ViTs.
\end{itemize}

\section{Background and Related Works}
\label{sec:background}
\subsection{The Functional Role of Spikes}
Prior studies associate the emergence of spikes to (1) \textit{Attention Sinks} \citep{AttentionSink} and (2) \textit{Value-state Drains} \citep{value-state-drain}, two phenomena that consistently co-occur at specific, low-semantic tokens across various input sequences. These phenomena are regarded as mechanisms that enable \textit{no-op} operations \citep{QuantizableTransformers}. By counteracting the sum-to-one constraint imposed by the softmax function \citep{AttentionIsOffByOne}, they prevent over-mixing and thus improve model generalization \citep{LLMsAttendFirstToken}. Fundamentally, spikes act as the upstream catalysts that drive these two effects \citep{active-dormant-heads}. By generating massive activation values at these specific tokens, they facilitate the formation of Attention Sinks and Value-state Drains \citep{MassiveActivations, SystematicOutliers}. Understanding these interconnected behaviors is crucial for advancing activation quantization \citep{Prefixquant}, long context streaming \citep{AttentionSink}, KV cache compression \citep{model-tells-you}, and many more. 


\subsection{The Bias Property of Spikes}
\label{sec:lifecycle}
These spikes follow predictable patterns: (1) they occur exclusively at specific tokens, termed \textit{sink tokens}, and (2) they exhibit a rigid lifecycle across model depth \citep{SystematicOutliers}. Specifically, they \textbf{emerge} in the MLPs of early layers, \textbf{persist} throughout the residual stream in the intermediate layers, and \textbf{disappear} in the final layers. Formalizing these characteristics, \citet{MassiveActivations} propose \textit{Bias Hypothesis}, which posits that the spike is essential for the functionality of the model, but carries no semantic meaning regarding the inputs. In these studies, the spike is interpreted as a high-level scalar bias within the residual stream, leaving the underlying geometric mechanics underexplored.


\subsection{Quantization Strategies}
The presence of massive spikes severely degrades quantization by stretching the dynamic range. Distinct from \textbf{Channel-wise Outliers}, which can be effectively mitigated via algorithmic approaches, \textbf{Massive Spikes} exhibit much larger magnitudes. These phenomena are fundamentally distinct \citep{MassiveActivations}; to avoid confusion, we reconcile the conflicting terminology found in prior literature in Appendix \ref{app:terminology}. It has been observed that prefixing specific tokens in the KV cache allows input tokens to attend to the prefix, preventing new spikes from arising within the input sequence \cite{CushionCache} and thereby enabling spike-free PTQ \cite{Prefixquant, ActivationSpikes}. This approach, however, relies on a heuristic that the spikes only occur in the first instance of specific low-semantic tokens. This method only works for modalities like language models where tokens are discrete, and does not readily generalize to other architectures like ViTs.
\begin{figure*}[t]
    \centering
    \includegraphics[width=1\linewidth]{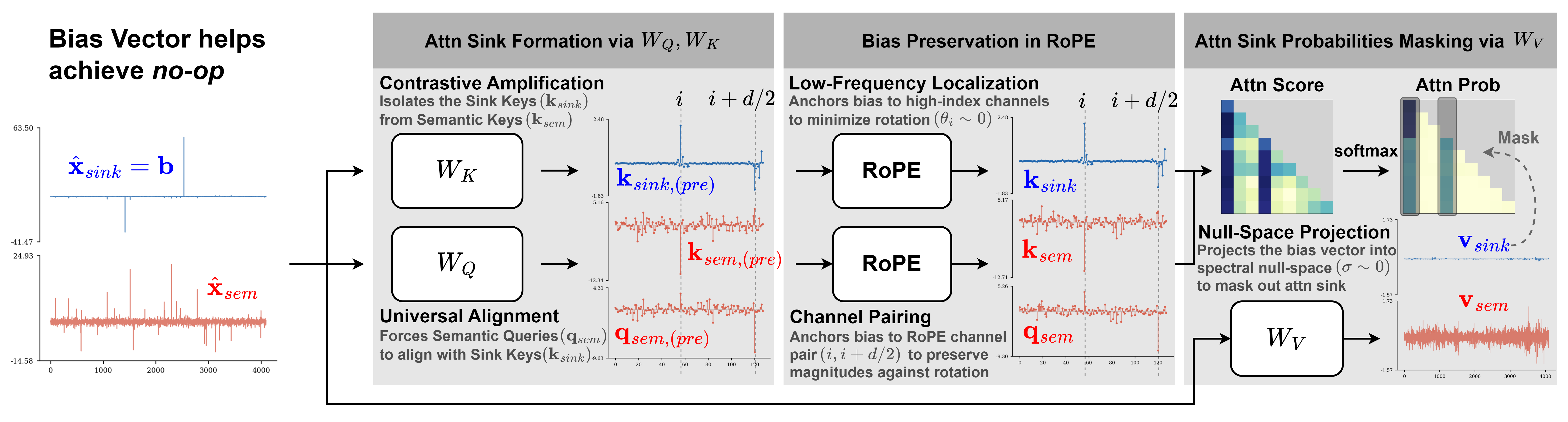}
    \caption{\textbf{Mechanistic Lifecycle of the Bias Vector in Self-Attention.} The rigid bias vector $\mathbf{b}$ orchestrates to allow no-update via three coordinated stages: (1) \emph{Attention Sink}: $W_K$ isolate this bias $\mathbf{b}$ from semantic tokens while $W_Q$ universally aligns queries toward it to create attention sinks; (2) \emph{Preservation under RoPE}: The bias $\mathbf{b}$ is encoded in zones of rotational stability (low-frequency bands and coherent pairs) to resist rotational perturbation; (3) \emph{Value-state Drains}: $W_V$ projects this bias $\mathbf{b}$ into a spectral null-space for masking out the attention sink probabilities}
    \label{fig:bias_vector_hypothesis}
\end{figure*}

\section{Bias Vector Hypothesis}
\label{sec:bias_vector_hypothesis}


While \citet{MassiveActivations} attribute the Massive Spikes phenomenon to high-level biases to enable \textit{no-op}, they primarily interpret these artifacts as scalar values observed at the residual stream. Concurrently, \citet{spike-sparse-sink} observe that these spikes interact with pre-normalization to form near-constant hidden representations, but treat them broadly as architectural "implicit parameters" without exploring their geometric mechanics.
In this work, we refine this understanding by proposing the \textit{Bias Vector Hypothesis}. Under this framework, we empirically verify the rigidity and bias property of the post-normalization vector (Section \ref{subsec:bias-vector-is-rigid}), geometrically analyze how this bias elicits attention sinks and value-state drains to enable no-op (Section \ref{sec:bias_mechanism}), and dissect how it achieves stability against token-dependent rotations in RoPE (Section \ref{sec:rope_interaction}).


\subsection{Formalizing the Bias Vector}
\label{subsec:bias-vector-is-rigid}
To pinpoint the bias vector, we analyze the \textbf{standardized vector} $\hat{\mathbf{x}}$ at the input of the self-attention block (post-normalization, pre-scaling in RMSNorm). Let $\mathbf{x} \in \mathbb{R}^d$ be the pre-norm activation of a token. We define the standardized vector as $\hat{\mathbf{x}} = \frac{\mathbf{x}}{\sqrt{\frac{1}{d} \|\mathbf{x}\|_2^2 + \epsilon}}$, where $\epsilon$ prevents division by zero.

Our hypothesis posits that for any sink token, this vector converges to a rigid \textbf{Bias Vector} $\mathbf{b}$ that is invariant across samples ($\hat{\mathbf{x}}_{\text{sink}} \to \mathbf{b}$). We define the bias pre-scaling to isolate the effect from the learnable scaling parameters ($\boldsymbol{\gamma}$), which act as layer-specific modulations that may vary across different normalization stages (e.g., pre-attention vs. post-attention).
In addition, we do not assume all sink tokens converge to a single global bias. Instead, we posit that the model maintains a discrete set of biases $\{\mathbf{b}_1, \dots, \mathbf{b}_K\}$, where each sink token aligns with a specific prototype $\mathbf{b}_k$. The number of biases $K$ is a model-specific characteristic \citep{WhenAttentionSinkEmerges}. We show the set of biases for LLaMA-2-7B and Mistral-7B-v0.3 in Appendix \ref{app:llama_template}.

As quantified in Figure \ref{fig:channelwise-variance}, the sink token of LLaMA-2-7B exhibits negligible channel-wise variance ($\sim 10^{-4}$), orders of magnitudes smaller than that of the semantic tokens ($\sim 10^0$). This extreme stability empirically corroborates our \textit{Bias Vector Hypothesis}. We further validate this functional role by replacing the dynamic sink token with its pre-computed static mean during inference. This replacement perfectly preserves downstream model performance (as detailed later in Table \ref{tab:LLM_concise_results}), inherently confirming that the vector carries no semantic meaning and acts purely as a structural bias.


\subsection{Bias Vector enables \textit{No-op} in Attention}
\label{sec:bias_mechanism}



The existence of the rigid bias vector $\mathbf{b}$ raises a critical question: how does this static input mechanistically enable the \textit{no-op} function? We advance the qualitative links drawn in prior work \citep{SystematicOutliers, MassiveActivations} by providing a quantitative geometric analysis. We reveal that the projection weights explicitly leverage this bias vector to induce \textit{attention sinks} and \textit{value-state drains} \citep{active-dormant-heads}, thereby enabling the \textit{no-op} mechanism, as shown in Figure \ref{fig:bias_vector_hypothesis}. Detailed experimental settings and analysis are provided in Appendix \ref{app:bias_mechanism}.

\paragraph{Bias Vector induces Attention Sinks.} 
The attention sink emerges from the direct geometric interaction between the rigid bias vector $\mathbf{b}$ and the Key and Query projection matrices: $W_K$ isolates the sink, while $W_Q$ aligns semantic tokens toward it.

\begin{figure*}[t]
    \centering
    \begin{subfigure}[b]{0.32\textwidth}
        \centering
        \includegraphics[width=\linewidth]{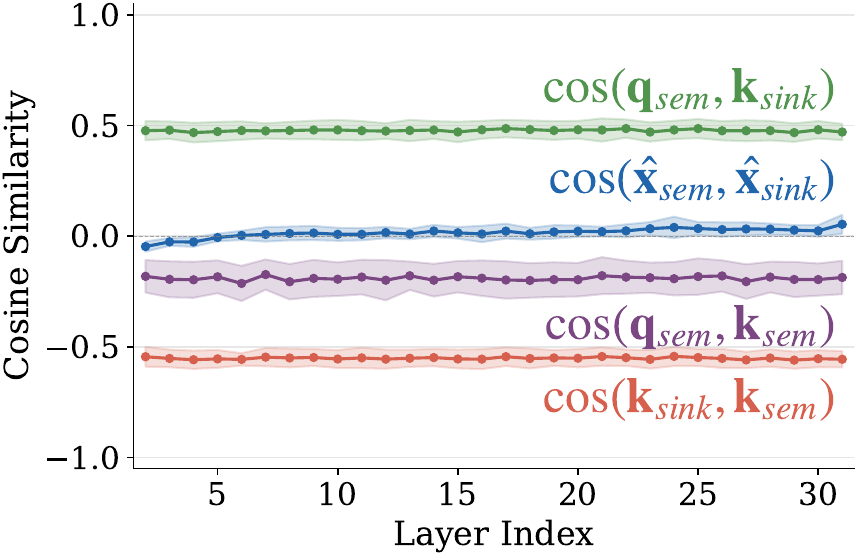}
        \caption{\textbf{$W_K, W_Q$ Alignment.} 
        }
        \label{fig:cosine_similarity}
    \end{subfigure}
    \hfill 
    \begin{subfigure}[b]{0.32\textwidth}
        \centering
        \includegraphics[width=\linewidth]{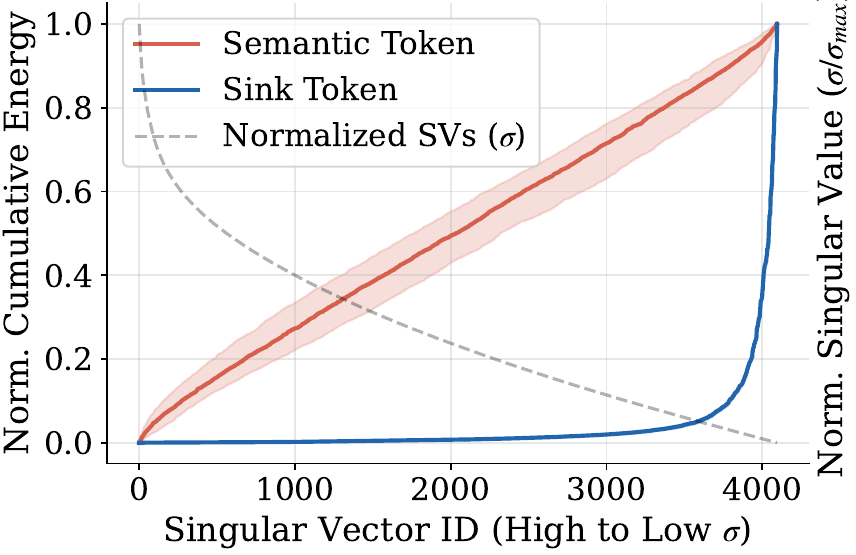}
        \caption{\textbf{$W_V$ Spectral Projection.} 
        }
        \label{fig:spectral_energy}
    \end{subfigure}
    \hfill
    \begin{subfigure}[b]{0.32\textwidth} 
        \centering
        \includegraphics[width=\linewidth]{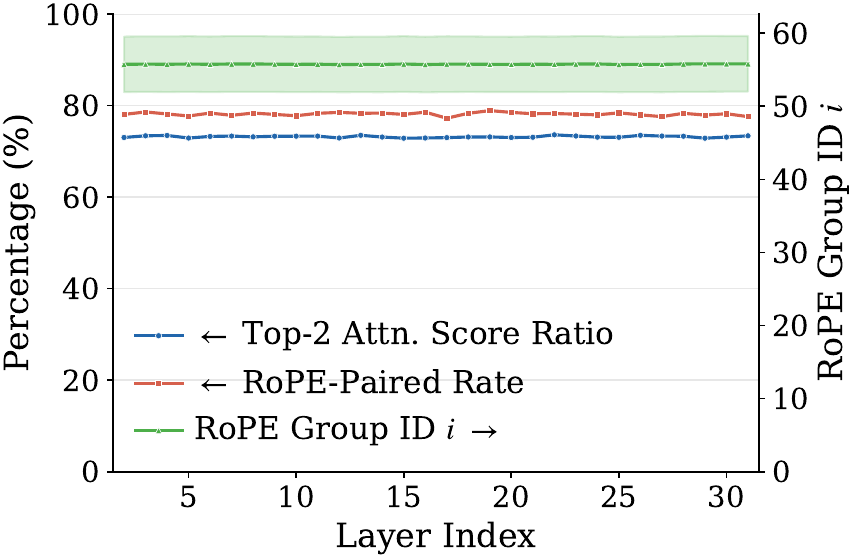} 
        \caption{\textbf{Stability through RoPE.}}
        \label{fig:rope_channel_analysis}
    \end{subfigure}
    
    \caption{\textbf{Geometric Mechanics of the Attention Sink.} (a) $W_K$ separates the sink token ($\mathbf{k}_{sink}$) from semantics, while $W_Q$ strictly aligns semantic queries ($\mathbf{q}_{sem}$) to the sink key. (b) Sink tokens (red) concentrate energy exclusively in the spectral tail of $W_V$ ($\sigma \approx 0$), unlike semantic tokens (blue) which span the amplification modes. (c) The channels driving the attention sink mechanism (top-2 score) strictly localize in low-frequency bands (green) and form coherent RoPE pairs (red) to minimize rotational variance via RoPE.}
    \label{fig:three_figures}
\end{figure*}

Specifically, the key projection ($W_K$) acts as a contrastive amplifier to structurally separate the sink token from semantic content. While the pre-projection cosine similarity between the standardized sink token $\hat{\mathbf{x}}_{sink}$ and a standard semantic token $\hat{\mathbf{x}}_{sem}$ is negligible ($\sim 0$), $W_K$ drives the similarity between their resulting keys ($\mathbf{k}_{sink}$ and $\mathbf{k}_{sem}$) down to approximately $-0.55$ (Figure \ref{fig:cosine_similarity}). In high-dimensional spaces (e.g., $d=4096$), this achieves extreme, near-antipodal separation.

Complementing this, $W_Q$ enforces a global attractor mechanism. It projects semantic queries ($\mathbf{q}_{sem}$) to strictly align with the isolated sink key $\mathbf{k}_{sink}$, maintaining a consistently positive similarity ($\sim +0.50$) that corroborates the ``static key'' observations of \citet{WhenAttentionSinkEmerges}. This universal alignment is remarkably stable across the persistence phase ($\sigma \approx 0.02$) and significantly exceeds baseline query-key interactions ($\sim -0.20$).

Collectively, these geometric dynamics structurally predispose all standard tokens to deposit their maximal attention scores onto the sink, empirically verifying how $W_K$ and $W_Q$ leverage the bias vector to construct the attention sink.

\paragraph{Bias Vector induces Value-state Drains.} 

Similarly, the value-state drain emerges from the geometric interaction between the bias vector $\mathbf{b}$ and the value projection matrix $W_V$. $W_V$ aligns $\mathbf{b}$ with its \textbf{spectral null-space}, generating a near-zero value vector to absorb the sink token's massive attention probability. We confirm this via spectral energy analysis (Figure \ref{fig:spectral_energy}), which demonstrates that the sink token concentrates its energy exclusively within the negligible tail components of $W_V$ ($\sigma_i \approx 0$). This stands in stark contrast to standard semantic tokens, whose energy is broadly distributed across the spectrum. Consequently, the projected value vector practically vanishes ($\mathbf{v}_{sink} \approx \mathbf{0}$), ensuring the sink token contributes no semantic update to the residual stream. This structural orthogonality not only explains prior observations \citep{value-state-drain, WhenAttentionSinkEmerges}, but also reinforces the active, functional role of the bias vector $\mathbf{b}$ in completing the \textit{no-op}.




\subsection{Active Preservation of Bias Under RoPE}
\label{sec:rope_interaction}


In Section \ref{sec:bias_mechanism}, we established that the bias vector $\mathbf{b}$ enforces a strict geometric alignment between $\mathbf{q}_{sem}$ and $\mathbf{k}_{sink}$ to induce attention sinks. In practice, Rotary Positional Embeddings (RoPE) \citep{Roformer} introduce token-dependent rotations that could disrupt this alignment across tokens, thereby potentially breaking the bias property. 

Specifically, RoPE partitions the $d$-dimensional representation into 2D subspaces and rotates the $i$-th subspace of a token at position $m$ by an angle $m\theta_i$ (where frequency $\theta_i = \Theta^{-2i/d}$). Consequently, the inner product between a semantic query at $m$ and a sink key at $n$ is modulated by a relative phase shift of $(m-n)\theta_i$. As $m$ varies across the sequence, this dynamic angular shift induces oscillations, destabilizing the static geometric alignment required for the \textit{no-op}. However, remarkably, our analysis reveals that the model actively resolves this geometric contradiction by localizing the structural bias within specific zones of rotational stability, providing evidence that significantly corroborate the \textit{Bias Vector Hypothesis}.

By analyzing the channel-wise contribution to the attention score across all layers (Figure \ref{fig:rope_channel_analysis}), we uncover the geometric strategies employed by the model to achieve this stability. First, we confirm the observation by \citet{SystematicOutliers} that the channel-wise attention contributions are highly \textbf{sparse} (Blue Line, Left Axis); the top-2 channels alone account for $\sim 75\%$ of the total positive attention score. Leveraging this extreme sparsity, the model isolate this attention sink feature from the destabilizing effect of RoPE effect using two specific spectral stabilization strategies:

\paragraph{Low-Frequency Bias. } As shown in Figure \ref{fig:rope_channel_analysis} (green line, right axis), the dominant sink features are disproportionately concentrated in high-index subspaces ($i \approx 55$ out of 64). Since the RoPE frequency $\theta_i$ decays exponentially with index $i$, this confirms the model targets the lowest-frequency band to minimize angular shift, effectively freezing the relative rotation.

\paragraph{Coherent Pairing. } As shown in Figure \ref{fig:rope_channel_analysis} (red line, left axis), in $\sim 80\%$ of cases, the top-2 channels form an exact RoPE 2D pair ($i, i+d/2$), indicating that the model coherently activates both dimensions of the rotational subspace. If the query alignment to the sink key relied only on a single channel $i$, the magnitude could oscillate sinusoidally as the energy rotated out of $i$ and into the RoPE paired channel $i+d/2$. By activating both elements, the model preserves magnitude robustness even as energy rotates with the 2D subspace.

The observed structural rigidity for rotation avoidance strongly corroborates the \textit{Bias Vector Hypothesis}. If the attention sink were merely a dynamic by-product of high-magnitude activations, it would not require such precise isolation from positional encoding. Instead, the model's explicit effort to shield these features in high-stability regions demonstrates that the bias vector is a \textbf{rigid, structural artifact} that lacks the degrees of freedom to adapt to large rotations. We detail the experiment settings, results, and interpretation in Appendix \ref{app:rope_analysis} and provide mathematical justifications for the findings in \ref{appx:rope-math}.

\subsection{Implications for Quantization}
Elevating the bias perspective to vector level is critical. Since the quantization bottlenecks stem from the rigid bias vector $\mathbf{b}$, we can decouple the spike-inducing sink tokens during the quantization process and reintroduce their bias effect later. This strategy eliminates the massive spikes that typically collapse quantization dynamic range without disrupting the model’s underlying \textit{no-op} mechanics. 
\section{\ours}
\label{sec:insertquant}

\begin{figure*}[t]
    \centering
    \includegraphics[width=\linewidth]{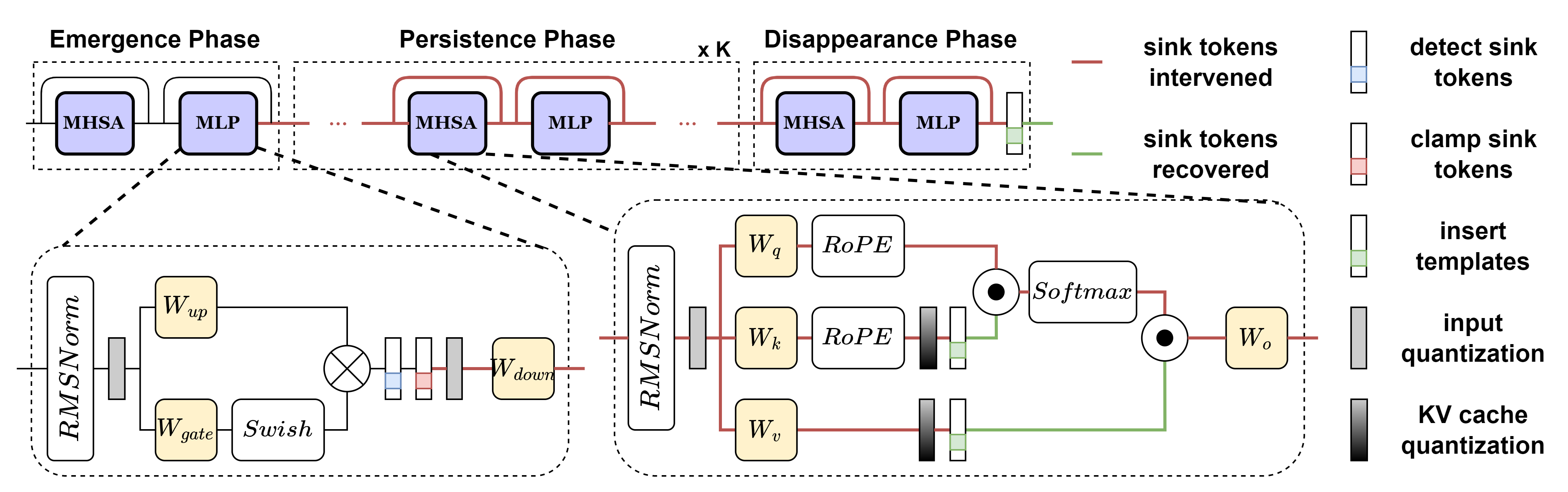}
    \caption{\textbf{Overview of the \ours Framework.} \ours eliminates massive spikes to enhance quantization through a three-step process. \textbf{Clamp Spikes}: Sink tokens carrying spikes are identified and clamped to zero in the Emergence Phase, as these tokens function as static bias terms. \textbf{Spike-free Quantization}: With spikes removed, the remaining activations become quantization-friendly. \textbf{Insert Bias Effects Back}: The effects of the clamped spikes are recovered in the subsequent layer by inserting pre-computed Key and Value template vectors.}
    \label{fig:insert_quant}
\end{figure*}


Leveraging the \textit{Bias Vector Hypothesis}, we propose \ours, a PTQ framework that separates the structural spikes from the content tokens. Since sink tokens converge to rigid structural biases rather than semantic content, they do not require runtime computation. Instead, \ours explicitly removes these tokens from the activation stream before quantization and restores their effect afterwards by inserting pre-computed vectors. 

Given that the sink token stabilizes into a known bias $\mathbf{b}$ at the RMSNorm output, we evaluate the propagation to see where clamping is safe. We observe that the inputs of $W_{MLP}$, $W_O$, and $Q$ are \textbf{functionally isolated}: the sink's computed MLP/Output states do not interact with semantic tokens, and crucially, its Query vector $\mathbf{q}_{sink}$ does not influence the attention scores of other tokens. Consequently, the input to these projections is redundant. In contrast, the inputs of $K$ and $V$ are \textbf{functionally critical}: because the global attention mechanism relies on token mixing, semantic tokens must attend to the sink's Key and Value vectors (where attention sinks and value-state drains formed). In Section \ref{sec:bias_mechanism}, we demonstrate that the bias $\mathbf{b}$ is structurally resilient to RoPE perturbation. Consequently, the transformation from $\mathbf{b}$ to $\mathbf{k}_{sink}$ is approximately deterministic, while the projection to $\mathbf{v}_{sink}$ is strictly deterministic. We refer to these two vectors as the \textbf{template vectors}.

\subsection{Calibration and Template Construction}
The process begins with an offline calibration phase to compute the template vectors of the sink tokens. 

\paragraph{Sink Token Detection and Template Collection.}
We first identify the indices of sink tokens at the spike emergence phases based on the magnitude statistics of the carried spikes, following the methodology of \citet{Prefixquant}. 
Once the positions are identified, we capture the precise vector values needed for insertion. For all subsequent layers, we collect the mean activation vectors of the detected tokens at $\mathbf{k}_{sink}$ and $\mathbf{v}_{sink}$ as templates.
We refer readers to Appendix \ref{app:insertquant_details} for detailed detection thresholds and template matching algorithms.

\subsection{Clamp and Insert}

The core insight of \ours is that the functional role of the sink tokens is \textit{deterministic}, meaning we do not need to compute them during runtime. This allows us to aggressively remove these tokens to benefit quantization, \textbf{knowing that we can reconstruct them later}. The process involves two steps applied to every layer in the persistence phase:

\begin{enumerate}
    \item \textbf{Clamp (Spike Elimination):} We clamp the activations of sink tokens to zero immediately after detection and before they enter the down projection. Once the sink tokens are clamped, they no longer carry the spikes throughout the network, thereby rendering the model quantization-friendly. 
    
    \item \textbf{Insert (Token Reconstruction):} We restore the functional role of the sink tokens by injecting pre-computed template vectors $\mathbf{k}_{sink}$ and $\mathbf{v}_{sink}$ directly at the input of the batched matrix multiplications (BMMs). This ensures that sink tokens retained valid targets for all tokens in the attention sink and value-state drain mechanism, despite the zeroed inputs.
\end{enumerate}

\subsection{Residual Restoration}


Finally, our clamping intervention disrupts the model's intrinsic \textit{cancellation mechanism} in the end of the network, where negative MLP spikes are meant to annihilate postive residual spikes \citep{SystematicOutliers}. To compensate for this, we perform a terminal restoration: we replace the sink tokens after the Disappearance Phase (Section \ref{sec:lifecycle}) at the residual stream with a pre-computed template derived from the calibration set. We confirm in Section \ref{sec:experiments} that the residual restoration, along with the template insertion, preserves model fidelity.

\section{Experiments}
\label{sec:experiments}
We evaluate \ours, which exploits \textbf{Bias Vector} to enable spike-free activation quantization. We first describe the experimental protocol and benchmarks for LLM quantization, then study generalization beyond text by quantizing ViT, where activations correspond to image tokens. Finally, we report system latency/memory overhead.

\subsection{Evaluation on Language Models}
\label{subsec:exp_setup}

\paragraph{Datasets.}
\ours collects the bias vector using a lightweight calibration set of 64 sequences uniformly sampled from the Pile dataset \citep{pile}, each truncated to 512 tokens. 

For evaluation of language models, we report zero-shot accuracy on eight common-sense reasoning (CSR) tasks (ARC-Easy, ARC-Challenge \citep{ARC}, BoolQ \citep{BoolQ}, PIQA \citep{PIQA}, SIQA \citep{SocialIQA}, HellaSwag \citep{HellaSwag}, OpenBookQA \citep{openbookQA}, and WinoGrande \citep{winogrande}) and perplexity on the WikiText-2 \citep{wikitext} test set using context lengths of 2048. All language evaluations use \texttt{lm-evaluation-harness} v0.4.2 \citep{eval-harness} with identical prompts and scoring across methods.

\paragraph{Models and baselines.}
We evaluate on LLaMA-3.2-3B \citep{llama3.2}, LLaMA-2-7B \citep{llama2}, LLaMA-2-70B, and Mistral-7B-v0.3 \citep{mistral7bv03}. For weight quantization, we adopt round-to-nearest (RTN) with fine-grained clipping threshold grid-search \citep{Prefixquant} and compare against three post-training activation quantization baselines: RTN, QuaRot-RTN (activation rotation via \citet{Quarot} with RTN weight quantization), and PrefixQuant~\cite{Prefixquant}. All methods use per-channel weight quantization, per-tensor static activation quantization, and group-wise dynamic KV-cache quantization with group size 128. We report results without any fine-tuning, isolating the effect of quantization and reflecting practical deployment settings where post-training quantization is preferred. Complete configurations for weights, activations, and the KV cache are provided in Appendix~\ref{app:exp_config}. We test several activation quantization methods under 4-bit precision setting and summarize the result in Table \ref{tab:LLM_concise_results}.

\paragraph{Results for Language Models.}
\begin{table}[t!]
\centering
\renewcommand{\arraystretch}{1.1}
\setlength{\tabcolsep}{6pt}
\footnotesize

\caption{\textbf{LLM quantization results.} We report (i) mean accuracy on eight zero-shot common-sense reasoning (CSR) tasks and (ii) WikiText-2 perplexity at context lengths 2048. QuaRot-RTN and PrefixQuant results are reproduced using the codebase of \citet{Prefixquant}.}

\resizebox{\columnwidth}{!}{%
\begin{tabular}{llcccc}
\toprule
\multirow{2}{*}{\textbf{Model}} & \multirow{2}{*}{\textbf{Method}} & \multicolumn{2}{c}{\textbf{CSR Accuracy ($\uparrow$)}} & \multicolumn{2}{c}{\textbf{WikiText-2 PPL ($\downarrow$)}} \\ \cmidrule(lr){3-4} \cmidrule(lr){5-6}
 & & \textbf{FP16} & \textbf{W4A4sKV4} & \textbf{FP16} & \textbf{W4A4sKV4} \\ \midrule

\multirow{4}{*}{LLaMA-3.2-3B} 
 & RTN            & 60.99 & 34.57 & 7.81 & 7126.79 \\
 & Quarot-RTN     & 61.04 & 38.42 & 7.81 & 59.68   \\
 & PrefixQuant    & 61.01 & 57.06 & 7.78 & 10.14   \\
 & InsertQuant    & 60.91 & 57.19 & 7.82 & 10.96   \\ \midrule

\multirow{4}{*}{LLaMA-2-7B} 
 & RTN            & 62.44 & 35.56 & 5.47 & 5859.52 \\
 & Quarot-RTN     & 62.44 & 34.45 & 5.47 & 424.73  \\
 & PrefixQuant    & 62.50 & 60.17 & 5.47 & 6.21    \\
 & InsertQuant    & 62.44 & 60.26 & 5.48 & 6.31    \\ \midrule

\multirow{4}{*}{LLaMA-2-70B} 
 & RTN            & 68.95 & 36.22 & 3.32 & 1352.96 \\
 & Quarot-RTN     & 68.91 & 53.49 & 3.32 & 9.64    \\
 & PrefixQuant    & 68.39 & 65.56 & 3.32 & 4.38    \\
 & InsertQuant    & 68.84 & 65.22 & 3.33 & 4.83    \\ \midrule

\multirow{4}{*}{Mistral-7B-v0.3} 
 & RTN            & 66.89 & 34.92 & 5.32 & 135012.38 \\
 & Quarot-RTN     & 66.92 & 40.18 & 5.32 & 55.91     \\
 & PrefixQuant    & 66.33 & 64.26 & 5.34 & 5.89      \\
 & InsertQuant    & 65.84 & 62.75 & 5.35 & 6.12      \\ \bottomrule

\end{tabular}%
} 
\label{tab:LLM_concise_results}
\end{table}
As shown in Table \ref{tab:LLM_concise_results}, \ours demonstrates robust performance across both full-precision and low-precision regimes. In the FP16 setting, \ours achieves CSR accuracy and WikiText-2 perplexity on par with the baseline, confirming that our surgical intervention on spikes based on the \textit{Bias Vector Hypothesis} preserves the model's functionality. Furthermore, in 4-bit weight/activation/KV cache static (W4A4sKV4) settings, our scheme significantly outperforms the RTN and rotation-based methods while closely matching the state-of-the-art performance of PrefixQuant. For instance, for LLaMA-2-7B, \ours yields a slightly higher CSR accuracy of 60.26 compared to PrefixQuant's 60.17, and maintains a Wikitext-2 perplexity of 6.21, closely tracking PrefixQuant's 6.31. 

We further show in the next section that \ours has better generalizability compared to PrefixQuant.
While PrefixQuant exploits the heuristic that spikes only appear in the first occurrence of specific tokens and mitigate them by prefixing a set of fixed discrete sink tokens into the KV cache, they are inherently limited to language models that possess discrete tokenizer definitions (e.g., ``\texttt{\textbackslash n}''). In contrast, \ours achieves the same functional stability by surgically injecting the sink's Key/Value templates. By replicating the sink mechanism geometrically rather than heuristically, we ensure generalizability to modalities where explicit sink tokens do not exist.

\subsection{Evaluation on ViTs}
Prior work has shown that ViTs~\citep{dosovitskiy2020image} used in CLIP~\citep{radford2021learning} and DINOv2~\citep{dinov2} exhibit similar massive spikes, where the spikes occur either in dedicated ``register tokens’’ or in irrelevant background tokens when the model does not use register tokens~\citep{MassiveActivations, darcet2023vision}. Because \ours imposes a stronger hypothesis that the continuous representation of a sink token converges to fixed bias vectors, we can flexibly extend \ours to quantizing ViTs, where existing prefix-based methods fail to expand to. 

The \ours implementation of ViT has two architecture-specific differences compared to that for the language models. First, as massive spikes in ViTs emerge in significantly deeper layers (e.g., Layer 12) compared to language models, we shift our spike detection to these later stages. Second, we omit the rotation transformation \citep{Quarot}, as the pre-LayerNorm structure breaks the required rotational invariance.

\paragraph{Setup.} 
We evaluate \ours on CLIP (ViT-L) and DINOv2 (ViT-L and ViT-B). For all ImageNet-1K (IN) \citep{russakovsky2015imagenet} evaluations, we randomly sample 1,000 images from the validation split. For CLIP, we perform zero-shot IN classification via normalized cosine similarity against text embeddings generated with the prompt ``a photo of a \{class\_name\}." We also evaluate zero-shot retrieval on the Flickr30k \citep{flickr30k} 1,000-image test split, reporting Recall@1 for both image-to-text (I2T) and text-to-image (T2I). For the vision-only DINOv2 models, we evaluate standard IN accuracy.


\begin{table}[t!]
\centering
\renewcommand{\arraystretch}{1.1}
\setlength{\tabcolsep}{4pt} 
\footnotesize

\caption{\textbf{ViT quantization results.} We report top-1 accuracy on ImageNet-1K (IN) and zero-shot image/text retrieval (Recall@1) on Flickr30k (T2I/I2T). By cleanly eliminating massive spikes, InsertQuant preserves model capacity and shows strong cross-modal generalizability across different architectures.}

\resizebox{\columnwidth}{!}{%
\begin{tabular}{clcccc}
\toprule
\multirow{2}{*}{\textbf{Bits}} & \multirow{2}{*}{\textbf{Method}} & \multicolumn{2}{c}{\textbf{CLIP ViT-L}} & \textbf{DINOv2 ViT-L} & \textbf{DINOv2 ViT-B} \\ \cmidrule(lr){3-4} \cmidrule(lr){5-5} \cmidrule(lr){6-6}
(W-A-KV) & & \textbf{IN Acc.} & \textbf{Flickr (T2I/I2T)} & \textbf{IN Acc.} & \textbf{IN Acc.} \\ \midrule

\multirow{2}{*}{16-16-16} 
 & Baseline       & 73.00 & 84.70 / 65.80 & 86.20 & 84.70 \\
 & InsertQuant    & 71.50 & 84.90 / 65.90 & 87.70 & 84.50 \\ \cdashline{1-6}
 
\multirow{2}{*}{8-8s-8} 
 & RTN            & 65.00 & 83.20 / 61.08 & 84.50 & 6.00  \\
 & InsertQuant    & 70.00 & 84.90 / 63.04 & 86.50 & 74.10 \\ \cdashline{1-6}

\multirow{2}{*}{4-8s-8} 
 & RTN            & 63.20 & 82.80 / 60.26 & 77.10 & 2.20  \\
 & InsertQuant    & 66.20 & 83.70 / 63.28 & 85.70 & 54.30 \\ \cdashline{1-6}

\multirow{2}{*}{4-6s-6} 
 & RTN            & 46.20 & 68.80 / 52.08 & 0.30  & 1.00  \\
 & InsertQuant    & 57.50 & 81.40 / 60.42 & 33.00 & 1.60  \\ \bottomrule

\end{tabular}%
} 
\label{tab:vit_compact_results}
\end{table}

\paragraph{Results for ViT.}
As shown in Table \ref{tab:vit_compact_results}, \ours successfully generalizes to vision architectures. Under FP16, replacing the dynamic sink tokens with their pre-calibrated static bias vectors preserves model capacity across all tasks. In low-precision settings, this mechanistic spike removal yields a highly quantization-friendly activation space, preventing the catastrophic accuracy degradation observed in the RTN baseline. For instance, in the W8A8sKV8 setting, RTN on DINOv2 ViT-B completely collapses to 6.00\% accuracy, whereas \ours preserves 74.10\%. These results demonstrate the cross-modal applicability of our framework and further corroborate the \textit{Bias Vector Hypothesis}.

\subsection{System Overhead of \ours}


The primary objective of \ours is not to introduce novel acceleration primitives, but to safely unlock the universal speed and memory benefits of ultra-low precision (e.g., W4A4) by preventing spike-induced accuracy collapse. Thus, rather than reporting total quantized memory savings—which are standard across all W4A4 methods—the critical system metric is the operational overhead introduced by our "clamp-and-insert" mechanism. To evaluate this, we measure the latency and memory overheads directly against the FP16 baseline. By implementing the spike detection and template insertion logic in Triton, \ours introduces a negligible latency overhead of just 4.69\% (measured on an NVIDIA A100 GPU with a 5-step warmup). Regarding memory, storing the necessary matching and insertion templates requires only 0.948 MB for a 13.5 GB LLaMA-2-7B model—an increase of merely 0.007\%. Ultimately, these minimal costs are fully justified by the complete elimination of massive spikes and the preservation of model fidelity.

\section{Conclusions}
In this work, we address the fundamental bottleneck of activation quantization by reframing massive spikes as structural vector-level features. Through the \textit{Bias Vector Hypothesis}, we established that spike-carrying tokens converge to rigid vector biases, which mechanistically interact with projection weights and RoPE to enable attention sink and value-state drain: the model utilizes the Key projection to create a high-magnitude contrastive attractor ($W_K$) while simultaneously using the Value projection ($W_V$) to nullify the output, effectively allowing a \textit{no-op} to be performed. Additionally, our analysis of RoPE reveals that models actively protect these biases in spectral \textit{stable zones}, confirming their role as rigid bias vectors. Capitalizing on these insights, we introduced \ours, a PTQ framework that surgically excises these vectors from the quantization loop. By explicitly clamping spikes and reinjecting their influence via pre-computed templates, we achieve a strictly spike-free activation space suitable for hardware-friendly per-tensor quantization. Our findings suggest that effective quantization requires a mechanistic understanding of how LLMs utilize spikes for enabling \textit{no-op}, paving the way for more efficient, quantized inference engines.

\section*{Limitations}
\paragraph{Architectural Scope and Applicability.} Our framework specifically targets models that naturally exhibit massive activation spikes. While we demonstrate strong generalizability across a diverse range of standard architectures (including LLaMA-2/3.2, Mistral, CLIP, and DINOv2), the exact manifestation of these spikes can vary based on underlying training dynamics. In addition, our method is unnecessary for models pre-trained with explicit architectural modifications designed to natively suppress these spikes \citep{gpt-oss,gatedattention}. Our core contribution remains a lightweight, post-training quantization solution for the vast majority of standard models burdened by spikes.

\paragraph{Theoretical Proof.} This work provides a mechanistic interpretation of massive spikes, analyzing their functional and geometric roles in pre-trained models. While we establish how these structural vectors execute the \textit{no-op} mechanism, we do not provide theoretical guarantees or optimization proofs explaining why the training process inherently converges to this specific geometric solution. Understanding the precise developmental trajectory of these bias vectors during pre-training remains a highly valuable direction for future theoretical research.

\section*{Ethics Statement}
This paper presents work whose goal is to advance the field of efficient
machine learning inference. The proposed quantization framework reduces
memory and compute requirements for deploying LLMs, which may contribute
to broader accessibility of large models. We do not foresee direct
harms specific to this work, though we acknowledge that more efficient
deployment of LLMs may amplify general societal impacts---both positive
and negative---associated with large-scale language models.

\bibliography{main}

\appendix
\section{Terminology of Activation Anomalies}
\label{app:terminology}

Relevant studies employ different terminologies to describe the high-magnitude activations in LLMs. To facilitate comparison, we map the terms used in recent literature to our taxonomy of \textbf{Channel-wise Outliers} (channel-wise vectors with moderate magnitude) and \textbf{Massive Spikes} (scalars with extreme magnitude).

\begin{table}[ht]
\centering
\caption{Terminology mapping for activation anomalies across recent literature.}
\label{tab:terminology_mapping}
\resizebox{\columnwidth}{!}{%
\begin{tabular}{@{}lll@{}}
\toprule
\textbf{Work}               & \textbf{Type 1 (Moderate, Consistent)} & \textbf{Type 2 (Extreme, Sparse)} \\ \midrule
\textbf{Ours}               & \textbf{Channel-wise Outlier}               & \textbf{Massive Spike}            \\
\citet{MassiveActivations}  & Outlier Feature                        & Massive Activation                \\
\citet{SystematicOutliers}  & Outlier                                & Outlier               \\
\citet{duquant}             & Normal Outlier                         & Massive Outlier                   \\
\citet{ActivationSpikes}    & Outlier                                & Activation Spike                  \\
\citet{Prefixquant}         & Channel-wise Outlier                   & Token-wise Outlier                \\
\citet{active-dormant-heads} & -                   & Residual-state Peak                \\ \bottomrule
\end{tabular}%
}
\end{table}

\section{Related Work for Attention Sinks}
\subsection{Massive Spikes and Attention Sink Mechanism}
 Prior studies attribute the emergence of such spikes to the probability constraint of the softmax operation in self-attention \cite{AttentionIsOffByOne}. In a multi-head residual-based transformer, each head is tasked with retrieving highly specific information; when the target information is absent, the head should yield zero contribution to the block output, effectively performing a \textit{no-op} to the previous representation to the residual stream. However, because softmax outputs are strictly positive and must sum to one, the mechanism cannot naturally produce a zero vector. To approximate this, the model designates specific tokens—termed \textbf{sink tokens}—to absorb the excess probability mass, thereby "sinking" the attention scores to zero out the contribution of semantic tokens \citep{AttentionSink}. Notably, \textit{attention sink} refers to not just the initial token, but all tokens that absorb disproportionate attention scores \citep{HiddenAttentionSink}. 

 Recent literature has formalized the geometric properties of attention sinks. \citet{spectralFilters} reveal that the spectral tails of the unembedding matrix are utilized to enable the attention sink mechanism, characterizing the non-semantic nature of the sink tokens. Focusing on the self-attention mechanism, \citet{WhenAttentionSinkEmerges} characterize sinks as learned ``key biases'' that consistently maintain high alignment with queries, coupled with value vectors of negligible $L_2$ norm to suppress residual updates. Complementing this, \citet{WhatAreYouSinking} posit that these sinks establish fixed reference frames to maintain geometric coordination against the rotational variance of positional encodings.


\section{Visualization of Bias Vectors}
\label{app:llama_template}

In Section \ref{sec:bias_vector_hypothesis}, we posited that sink tokens align with a discrete set of prototypical bias vectors $\{\mathbf{b}_1, \dots, \mathbf{b}_K\}$. Here, we visualize the learned bias templates for LLaMA2-7B and Mistral-7B-v0.3 at the RMSNorm output.

\begin{figure*}[ht]
    \centering
    \begin{subfigure}[b]{0.38\textwidth}
        \centering
        \includegraphics[height=4.0cm, width=\linewidth, keepaspectratio]{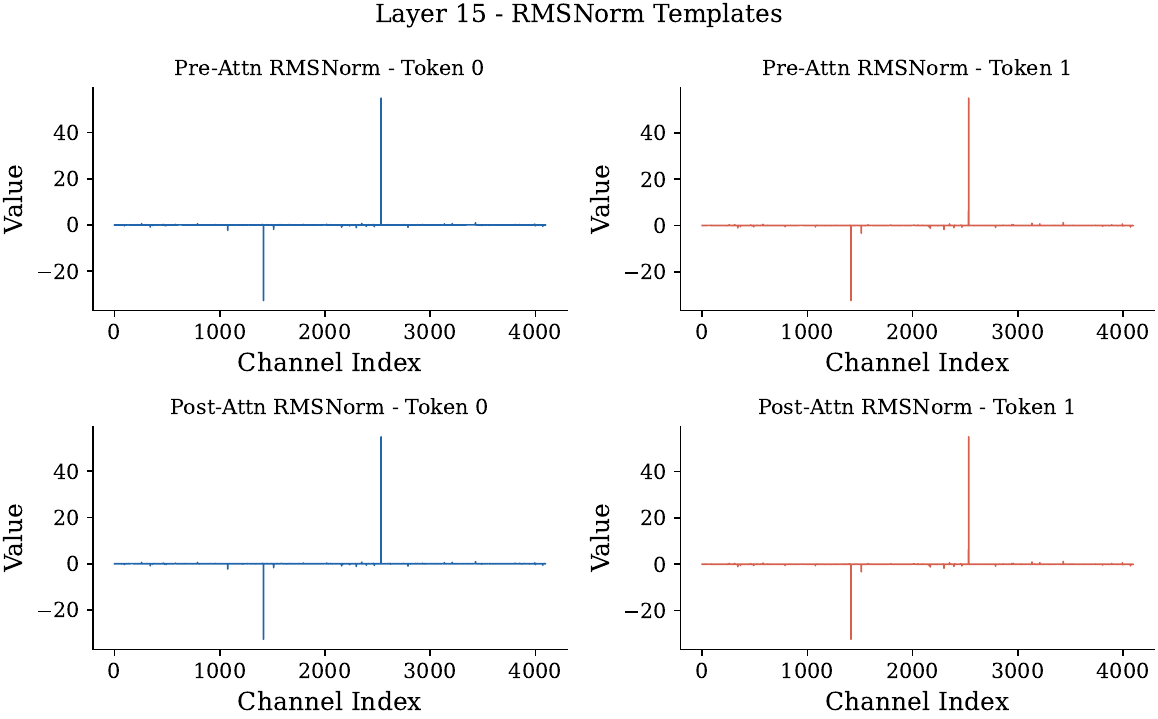}
        \caption{\textbf{LLaMA2-7B ($K=2$)}}
        \label{fig:llama_bias}
    \end{subfigure}
    \hfill 
    \begin{subfigure}[b]{0.58\textwidth}
        \centering
        \includegraphics[height=4.0cm, width=\linewidth, keepaspectratio]{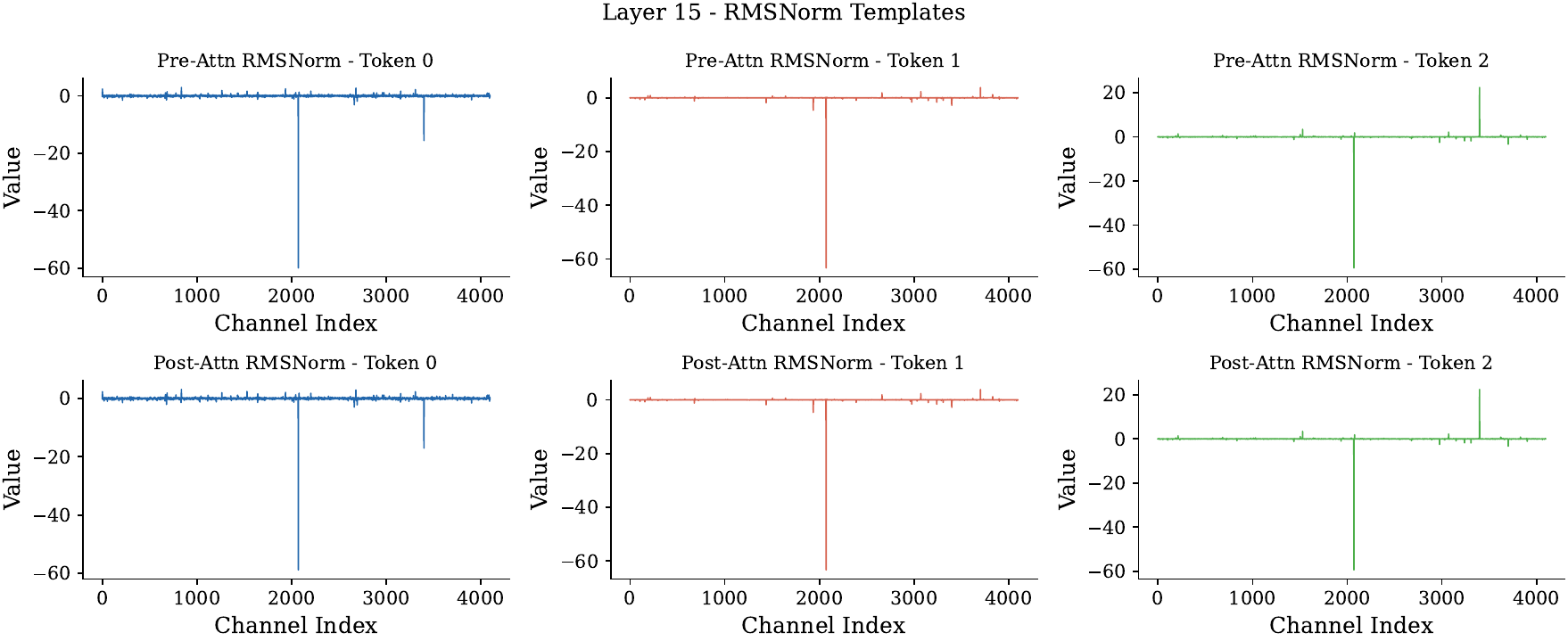}
        \caption{\textbf{Mistral-7B-v0.3 ($K=3$)}}
        \label{fig:mistral_bias}
    \end{subfigure}
    
    \caption{\textbf{Visualization of Bias Templates.} Comparison of the learned bias prototypes ($\mathbf{b}_k$) at the RMSNorm output. (a) The two LLaMA2 templates are nearly identical, overlapping in both spike location and magnitude. (b) The three Mistral templates share a global structure (e.g., spike at $\sim$ 2100) but exhibit visible variations in shape and secondary features.}
    \label{fig:bias_vectors}
\end{figure*}

\textbf{LLaMA2-7B ($K=2$):} As shown in Figure \ref{fig:llama_bias}, the two identified bias templates exhibit a \textbf{remarkable degree of similarity}. Both vectors are dominated by a negative spike around channel 1400 and a positive spike around channel 2600. The magnitude and direction of these features are nearly identical across the two templates, suggesting that LLaMA2-7B utilizes a highly consistent bias structure across different sink tokens.

\textbf{Mistral-7B-v0.3 ($K=3$):} As shown in Figure \ref{fig:mistral_bias}, Mistral-7B maintains three distinct sink templates. While all three share a similar global structure---characterized by a dominant negative spike near channel 2100---they exhibit \textbf{visible variations}. Specifically, the secondary features and exact magnitudes differ between the prototypes (e.g., Template 2 vs Template 0), indicating that the model preserves fine-grained distinctions between different sink token types rather than collapsing them into a single vector.

\section{Channel-wise Variance Analysis Across Layers}
\label{app:variance_analysis}

In this section, we analyze the channel-wise variance across LLaMA2-7B to track the persistence of the \textbf{Bias Vectors} from the emergence phase to disappearance phase.

\subsection{Methodology}
We compute the channel-wise variance of activation magnitudes across 64 samples from the \textbf{Pile} calibration set. The sink token detection method is described in Appendix \ref{app:insertquant_details}.

\subsection{Layer-wise Observations}
\begin{figure*}[t]
    \centering
    \includegraphics[width=\linewidth]{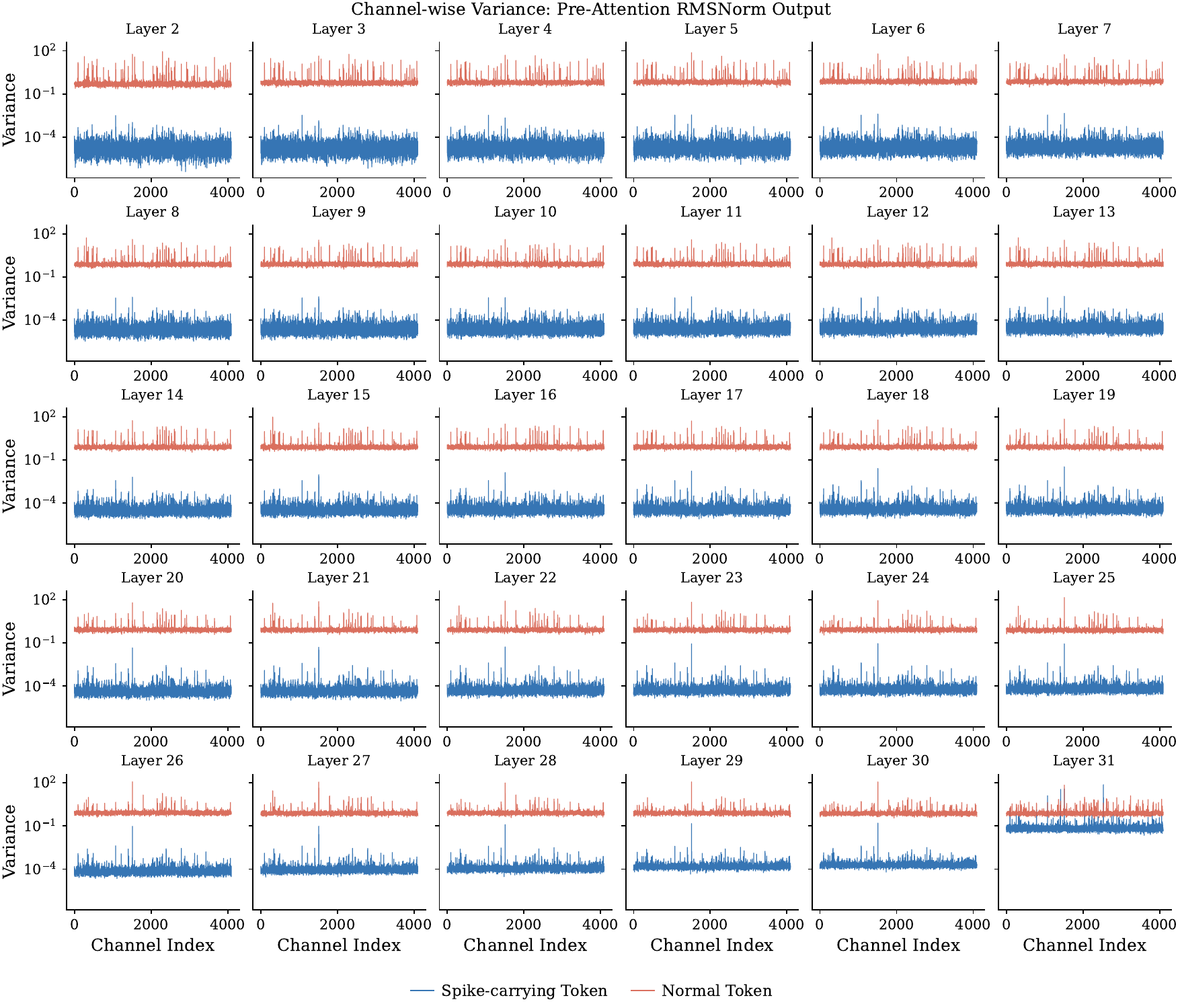}
    \caption{\textbf{Channel-wise variance at the Pre-Attention RMSNorm output across layers.} The sink tokens (blue) exhibit extremely low variance compared to normal semantic tokens (orange) until the disappearance phase at Layer 31. The learnable scaling factor is merged to the subsequent weights. Here for sink tokens, we show the second bias vector $\mathbf{b}_2$ of LLaMA2-7B, which has slightly larger variances compared to $\mathbf{b}_1$}
    \label{fig:all_layers_pre_attn_variance}
\end{figure*}

\begin{figure*}[t]
    \centering
    \includegraphics[width=\linewidth]{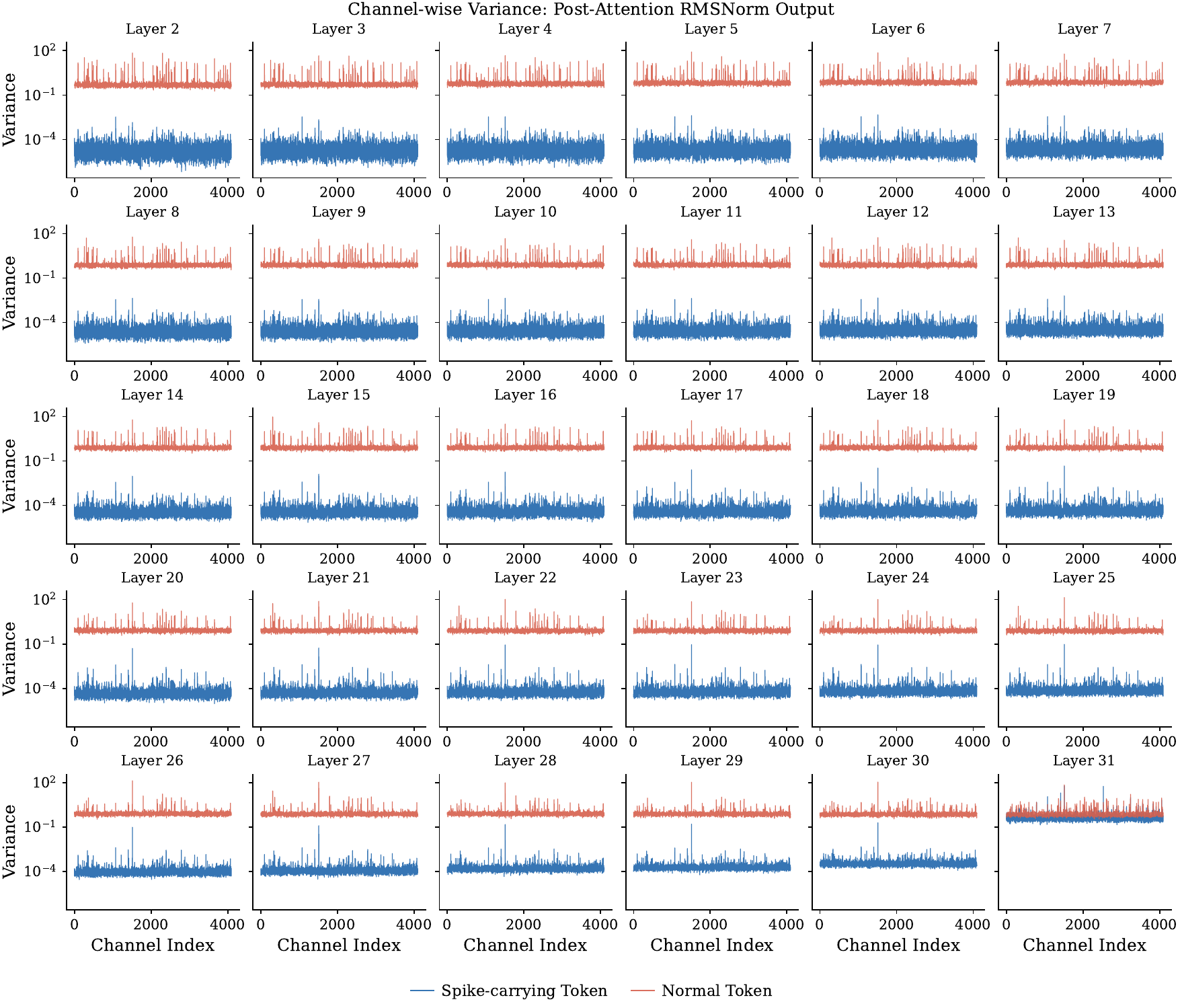}
    \caption{\textbf{Channel-wise variance at the Post-Attention RMSNorm output across layers.} The trend remains consistent with the pre-attention observations, confirming the stability of the template vector through the attention block. The variance for sink tokens rise significantly since the disappearance phase at Layer 31. }
    \label{fig:all_layers_post_attn_variance}
\end{figure*}

Figures \ref{fig:all_layers_pre_attn_variance} and \ref{fig:all_layers_post_attn_variance} visualize the channel-wise variance at the pre-scaling output of the RMSNorm (both pre- and post-attention) for all applicable layers.

\begin{itemize}
    \item \textbf{Consistency (Layers 2--30):} The channel-wise variances for spike-carrying tokens (blue curves) remain consistently negligible ($< 10^{-4}$) compared to normal tokens (orange curves) from Layer 2 through Layer 30. This confirms that the ``Sink Template'' is not a local artifact but a rigid, global feature vector preserved across the network's depth.
    
    \item \textbf{Start Point (Layer 2):} We initiate our observation at Layer 2, as the detection of emergent spike-carrying tokens occurs at the input of the down-projection layer in the first MLP block (Layer 1), as noted in the lifecycle analysis \citep{SystematicOutliers}.
    
    \item \textbf{Disappearance (Layer 31):} A distinct phase transition is observed at Layer 31, where the variance of the spike-carrying tokens abruptly increases to match that of normal tokens. This corroborates the ``Disappearance'' phase described in Section \ref{sec:lifecycle}: the massive spikes are explicitly canceled out by the MLP in Layer 30. Consequently, by Layer 31, these tokens no longer carry the rigid sink signal and return to a standard semantic variance profile.
\end{itemize}

\section{Geometric and Spectral Characterization of Attention Sinks}
\label{app:bias_mechanism}

In this section, we provide the detailed experimental setup and mathematical formulation used to derive the mechanistic analysis of attention sinks presented in Section \ref{sec:bias_mechanism}. To analyze the direct interaction between the bias vectors and the weights, we fuse $\boldsymbol{\gamma}$ of the RMSNorm layer into the projection matrices following the method described in \citet{Quarot}.

\subsection{Data Sampling and Preparation}

To ensure a robust analysis of the interaction between activation patterns and projection weights, we utilized a calibration set of 64 sequences randomly sampled from the \textsc{Pile} dataset, each with a sequence length of 512 tokens.

For each layer in the persistence phase (Layers 2--30), we classified tokens into two categories based on sink token detection and template matching:
\begin{itemize}
    \item \textbf{Sink Tokens  ($\hat{\mathbf{x}}_{sink} \sim \mathbf{b}$):} Randomly sampled from positions identified as sink tokens.
    \item \textbf{Normal Semantic Tokens ($\hat{\mathbf{x}}_{sem}$):} Randomly sampled from the remaining non-spike positions to represent standard semantic content.
\end{itemize}
All geometric analyses were performed using the effective projection matrices (with fused RMSNorm scaling factors) to strictly model the interaction between the template and the attention mechanism.

\subsection{Geometric Alignment Analysis}

\begin{figure*}[t]
    \centering
    \includegraphics[width=\linewidth]{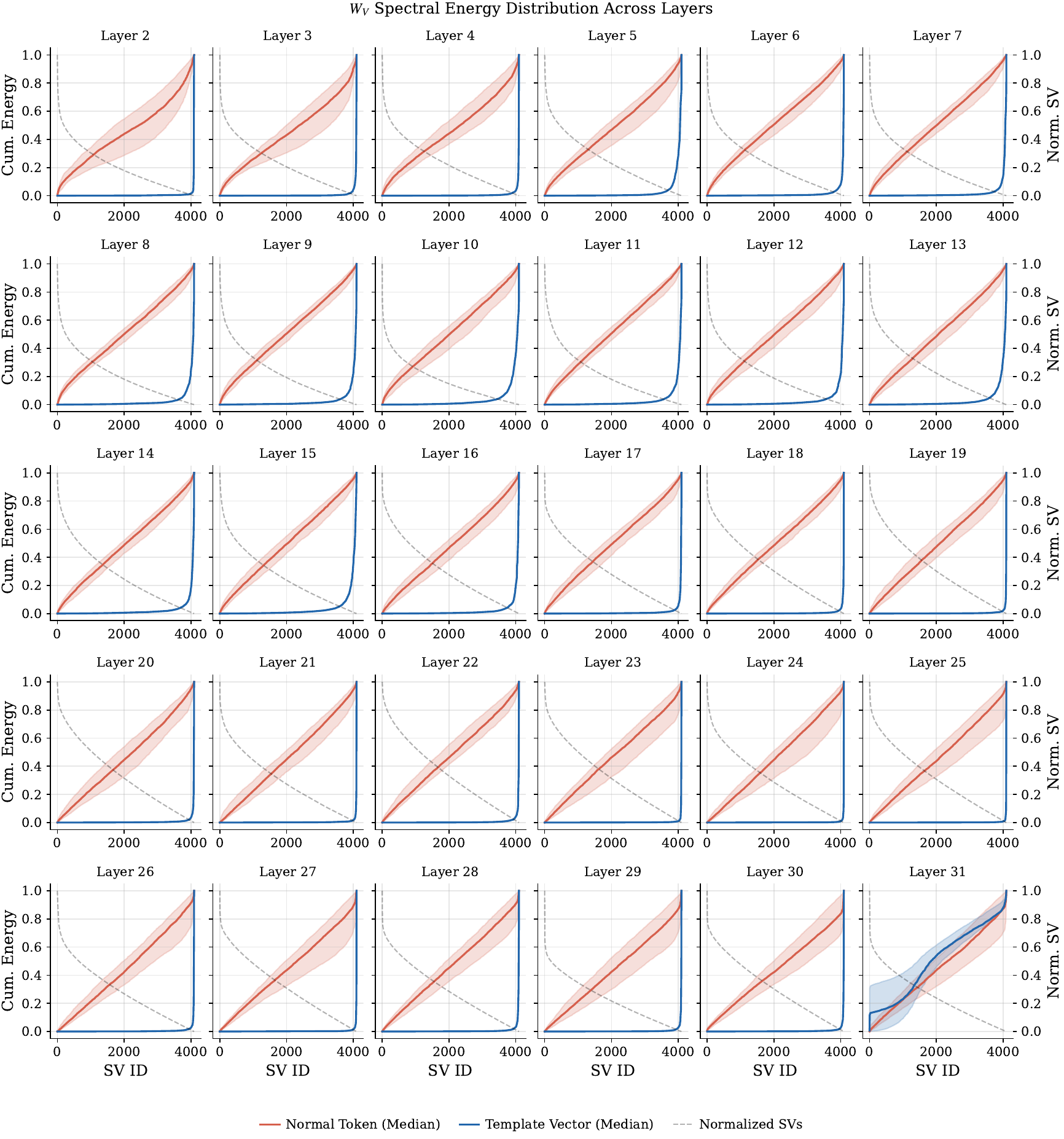}
    \caption{\textbf{Layer-wise Consistency of Spectral Null-space Projection.} Analysis of the Cumulative Spectral Energy (CSE) for Normal Tokens (Blue) and Template Vectors (Red) with 5th--95th percentile bands across all layers in the persistence phase (Layers 2--30). The Grey dashed line indicates the normalized singular value spectrum of $W_V$. The pattern is strictly consistent across the entire network depth: normal tokens distribute energy broadly across the semantic subspace (high $\sigma$), while template vectors effectively exist only in the null-space tail (negligible $\sigma$). This confirms that the informational suppression mechanism ($W_V$ acting as a Gatekeeper) is a global, rigid structural property of the model. Entering the disappearance phase (Layer 31), $W_V$ no longer possesses this spectral filtering property.}
    \label{fig:all_layers_spectral}
\end{figure*}

To generate the cosine similarity trajectories shown in Figure \ref{fig:cosine_similarity}, we computed the cosine similarity for the interaction pairs (e.g., $W_Q \hat{\mathbf{x}}_{sem} \leftrightarrow W_K \mathbf{b}$) at every layer. The solid lines represent the mean similarity across the sampled sequences, while the shaded regions represent the standard deviation ($\pm 1\sigma$). This metric quantifies the structural "Push-Pull" geometry, confirming that the model optimizes $W_Q$ and $W_K$ to enforce a high attention score for the sink template regardless of the input semantics.


\subsection{Spectral Energy Analysis}
To rigorously verify that the bias vector $\mathbf{b}$ drives the \textbf{value-state drain}---specifically, that $W_V$ actively nullifies this bias signal to enable the \textit{no-op} mechanism---we perform a spectral analysis using Singular Value Decomposition (SVD).

\textbf{Decomposition.} We decompose the value projection matrix $W_V \in \mathbb{R}^{d \times d}$ into $W_V = U \Sigma V^\top$, where the columns of $V$ form the orthonormal basis of the input space sorted by singular value magnitude ($\sigma_1 \ge \dots \ge 0$).

\textbf{Cumulative Spectral Energy (CSE).} For a normalized input token $\mathbf{z}$, we project it onto the right singular vectors $V$ to obtain the spectral coefficients $\mathbf{c} = \mathbf{z} V$. The Cumulative Spectral Energy is calculated as:
\begin{equation}
    \text{CSE}_\mathbf{z}(k) = \sum_{i=1}^{k} (\mathbf{z} \cdot \mathbf{v}_i)^2
\end{equation}
In Figure \ref{fig:spectral_energy}, we report the median CSE curve (50th percentile) aggregated over the sampled tokens, with the shaded bands representing the 5th and 95th percentiles.

\textbf{Consistency and Mechanism.} We further provide the spectral energy distribution for all layers in the persistence phase (introduced in Section \ref{sec:lifecycle}) in Figure \ref{fig:all_layers_spectral}. The results demonstrate that this trend is consistent across all layers where massive spikes are present: semantic tokens distribute energy broadly (linear trend), while template tokens concentrate energy exclusively in the null-space tail. This spectral orthogonality elucidates the physical implementation of the \textbf{no-update} mechanism: by confining the template vector to the null-space of $W_V$, the model ensures that even when the attention mechanism assigns a maximal weight to the sink token, the resulting update vector added to the residual stream has negligible magnitude, effectively preserving the hidden state. 
\section{Analysis of RoPE Interaction and Rotational Stability}
\label{app:rope_analysis}


In this section, we provide a detailed analysis of how the attention sink mechanism interacts with Rotary Positional Embeddings (RoPE). We first clarify the specific RoPE implementation used in LLaMA2-7B architectures, then detail the methodology for computing channel contribution scores, and finally present granular evidence of channel pairing.

\subsection{RoPE Mechanics and Implementation Differences}
While the original RoPE work \citep{Roformer} defines rotation by pairing adjacent feature channels $(2i, 2i+1)$, the LLaMA2-7B architecture (and its HuggingFace implementation) employs a different pairing strategy that is critical for interpreting our results.

In LLaMA2-7B, the head dimension $d$ is split into two halves. A feature at index $i$ (where $0 \le i < d/2$) is paired with the feature at index $i + d/2$. For a given position $m$, the rotation is applied to this pair vector $\mathbf{v} = [x_i, x_{i+d/2}]^\top$ via the matrix:
\begin{footnotesize}
    \begin{equation}
        \begin{pmatrix} x'_i \\ x'_{i+d/2} \end{pmatrix} = 
        \begin{pmatrix} 
        \cos (m\theta_i) & -\sin (m\theta_i) \\ 
        \sin (m\theta_i) & \cos (m\theta_i) 
        \end{pmatrix} 
        \begin{pmatrix} x_i \\ x_{i+d/2} \end{pmatrix}
    \end{equation}
\end{footnotesize}
where the frequency $\theta_i = b^{-2i/d}$ decays exponentially as $i$ increases. Crucially, this implies that:
\begin{itemize}
    \item \textbf{Frequency Distribution:} The lowest frequencies (slowest rotation) correspond to the largest indices $i \approx d/2$.
    \item \textbf{Pairing Distance:} Rotational invariance requires utilizing the 2D subspace formed by indices separated by exactly $d/2$ (i.e., $|c_1 - c_2| = 64$ for $d=128$).
\end{itemize}
This implementation context explains why our analysis focuses on "modulo 64" indices and pairing distances of 64.

\subsection{Methodology: Channel Contribution Analysis}
To identify the specific features driving the attention sink, we move beyond simple magnitude analysis (L2 norm) and focus on the \textbf{interaction} between queries and keys. The attention score is the dot product $q \cdot k = \sum_c q_c k_c$. A channel contributes to the sink mechanism only if it has a large positive product $q_c k_c$ (alignment).

We compute the \textbf{Channel Contribution Score} vector $\mathbf{c}$ for each head using randomly sampled tokens:
\begin{equation}
    \mathbf{c} = \mathbf{q}_{sem} \odot \mathbf{k}_{sink} \in \mathbb{R}^d
\end{equation}
where $\odot$ denotes the element-wise product, $\mathbf{q}_{sem}$ is a randomly sampled semantic query vector, and $\mathbf{k}_{sink}$ is a randomly sampled sink key vector.

\textbf{Positive Contribution Ratio.} To quantify sparsity, we calculate the proportion of the total attractive force generated by the top-2 channels. We normalize by the sum of \textit{positive} contributions only:
\begin{equation}
    R = \frac{\sum_{j \in \text{Top-2}} \mathbf{c}_j}{\sum_{k: \mathbf{c}_k > 0} \mathbf{c}_k}
\end{equation}
We strictly filter for positive contributions because the attention sink mechanism is linked with the generation of large \textit{positive} attention scores (attractors). Misaligned (negative) channels represent "friction" rather than the driving force of the sink, and including them would cause numerical instability when they cancel out positive terms in the denominator. This metric ($R$) thus represents the percentage of the attention sink's "spark" that is attributable to the single dominant feature pair.

\subsection{Mechanistic Interpretation of Channel Pairing}
Our analysis reveals that in $\sim 80\%$ of attention heads, the top-2 channels form an exact RoPE pair ($i, i+d/2$). This redundancy provides a critical geometric property: \textbf{Signal Stability in the 2D Subspace.}

Consider a feature encoded in a single channel $i$. As the token position $m$ increases, RoPE rotates the energy of this feature continuously between channel $i$ and its pair $i+d/2$.
\begin{itemize}
    \item \textbf{Single Channel Instability:} If the query $Q$ only attended to channel $i$, the resulting dot product would oscillate sinusoidally as the energy rotated out of $i$ and into the ignored channel $i+d/2$. This would cause the attention sink to "flicker" on and off depending on the token position.
    \item \textbf{Paired Channel Stability:} By allocating significant attention mass to \textit{both} components of the pair ($i$ and $i+d/2$), the model attends to the \textbf{magnitude} of the 2D vector rather than its projection onto a single axis. Since RoPE acts as a rotation, it preserves the Euclidean norm $\|\mathbf{v}\|_2$ of the 2D vector.
\end{itemize}
It is important to note that this does not guarantee perfect rotational invariance, as the contribution scores for the two channels ($\mathbf{c}_i$ and $\mathbf{c}_{i+d/2}$) may not be identical. However, by activating both dimensions, the model ensures that a substantial portion of the signal magnitude remains accessible regardless of the rotation angle, making the attractor signal significantly more stable than a single-channel projection.

\subsection{Granular Analysis of Channel Pairing}
To validate this at a granular level, Table \ref{tab:rope_layer15} details the channel statistics for a representative layer (Layer 15) of LLaMA2-7B. The data confirms that the pairing is precise and systematic. For instance, Head 0 utilizes channels 118 and 54 ($118-54=64$), and Head 23 relies on 117 and 53 ($117-53=64$). In cases where pairing is not exact (e.g., Head 1), the score ratio is typically lower ($58.4\%$), suggesting these heads may be less specialized for the sink function or rely on different mechanisms.


\subsection{Conclusion: Structural Rigidity in Dynamic Space}
These distinct localization phenomena—temporal grounding at the sequence start and spectral concentration in low-frequency, paired channels—converge on a single mechanistic purpose: mitigating the dynamic perturbation of RoPE to preserve a static signal. This implicit avoidance of rotation strongly corroborates the \textbf{Bias Vector Hypothesis}. If the attention sink were merely a dynamic by-product of high-magnitude activations, it would not require such precise isolation from positional encoding. Instead, the model's explicit effort to shield these features in ``rotational stability zones'' demonstrates that the bias vector is a \textbf{rigid, structural artifact} that lacks the degrees of freedom to adapt to rotation.

\begin{table}[t]
\centering
\tiny
\caption{RoPE Channel Analysis for Layer 15}
\label{tab:rope_layer15}
\resizebox{\columnwidth}{!}{%
\begin{tabular}{ccccc}
\toprule
Head ID & Attn Score Ratio (\%) & Top 1 Channel & Top 2 Channel & Paired? \\
\midrule
0 & 75.7 & 118 & 54 & \checkmark \\
1 & 58.4 & 52 & 63 &  \\
2 & 66.3 & 115 & 51 & \checkmark \\
3 & 80.3 & 60 & 124 & \checkmark \\
4 & 74.4 & 54 & 118 & \checkmark \\
5 & 45.7 & 60 & 59 &  \\
6 & 74.2 & 53 & 117 & \checkmark \\
7 & 52.5 & 61 & 51 &  \\
8 & 76.3 & 116 & 52 & \checkmark \\
9 & 68.7 & 115 & 51 & \checkmark \\
10 & 88.2 & 55 & 119 & \checkmark \\
11 & 60.8 & 117 & 53 & \checkmark \\
12 & 47.3 & 58 & 124 &  \\
13 & 65.2 & 126 & 62 & \checkmark \\
14 & 64.3 & 119 & 124 &  \\
15 & 50.8 & 117 & 53 & \checkmark \\
16 & 53.1 & 123 & 61 &  \\
17 & 73.3 & 119 & 55 & \checkmark \\
18 & 85.6 & 54 & 118 & \checkmark \\
19 & 81.5 & 61 & 125 & \checkmark \\
20 & 86.5 & 117 & 53 & \checkmark \\
21 & 84.6 & 120 & 56 & \checkmark \\
22 & 57.9 & 116 & 52 & \checkmark \\
23 & 91.7 & 117 & 53 & \checkmark \\
24 & 80.7 & 118 & 54 & \checkmark \\
25 & 90.4 & 58 & 122 & \checkmark \\
26 & 77.0 & 55 & 119 & \checkmark \\
27 & 56.4 & 59 & 63 &  \\
28 & 71.8 & 61 & 125 & \checkmark \\
29 & 68.5 & 62 & 116 &  \\
30 & 87.2 & 63 & 127 & \checkmark \\
31 & 92.3 & 53 & 117 & \checkmark \\
\bottomrule
\end{tabular}%
}
\end{table}

\section{Mathematical Justification of Structural Bias under RoPE}
\label{appx:rope-math}

In Section \ref{sec:rope_interaction}, we empirically demonstrated two critical phenomena regarding attention sinks: (1) the structural bias reliably localize within high-index channel pairs, and (2) the bias exhibit coherent activation, simultaneously occupying both channels of their corresponding 2D subspace. This appendix provides the mathematical justification for these observations, proving that such positioning and pairing are mathematically required to maintain a stable, rotation-invariant global attractor under RoPE.

\subsection{Formulating the Structural Bias under RoPE}

The core premise is that the massive spikes after RMSNorm act as rigid, structural \textbf{bias vectors} that enable a \textit{no-op} attention mechanism. Their function is to create a structurally stable attention sink for subsequent semantic queries, regardless of their relative sequence position.

In models utilizing RoPE, the hidden dimension $d$ is partitioned into $d/2$ independent 2D subspaces. Assuming the standard \texttt{rotate\_half} implementation, the query and key sub-vectors for a pair index $i \in \{0, 1, \dots, d/2 - 1\}$ are defined as:
$$\mathbf{q}_i = \begin{pmatrix} q_i \\ q_{i+d/2} \end{pmatrix}, \quad \mathbf{k}_i = \begin{pmatrix} k_i \\ k_{i+d/2} \end{pmatrix}$$

We hypothesize that the model learns to embed its massive structural bias into a specific subset of these subspaces, formally denoted as $\mathcal{I}_{bias} \subseteq \{0, 1, \dots, d/2 - 1\}$. To understand how the network protects this signal from being degraded by positional encoding, we isolate our analysis to the components within $\mathcal{I}_{bias}$.

Before positional rotation is applied, the base attention score contributed by this bias is the sum of the inner products across $\mathcal{I}_{bias}$. Expressing this geometrically—where $\alpha_i$ is the initial angle between $\mathbf{q}_i$ and $\mathbf{k}_i$—yields:
\begin{footnotesize}
    \begin{equation}
        S_{pre, bias} = \sum_{i \in \mathcal{I}_{bias}} \|\mathbf{q}_i\|_2 \|\mathbf{k}_i\|_2 \cos(\alpha_i)
    \end{equation}
\end{footnotesize}

RoPE applies a 2D rotation matrix $R_i(\Delta m)$ to each subspace, parameterized by the relative sequence distance $\Delta m = m - n$ and the channel-specific frequency $\theta_i = b^{-2i/d}$. Because this rotation shifts the relative angle between the 2D vectors by $\Delta m \cdot \theta_i$, the post-RoPE bias contribution becomes:
\begin{footnotesize}
    \begin{equation}
        S_{post, bias}(\Delta m) = \sum_{i \in \mathcal{I}_{bias}} \|\mathbf{q}_i\|_2 \|\mathbf{k}_i\|_2 \cos(\alpha_i - \Delta m \cdot \theta_i)
    \end{equation}
\end{footnotesize}

For the structural bias to reliably serve as an attention sink, $S_{post, bias}(\Delta m)$ must remain massively positive and robust across the entire context window. In the following subsections, we demonstrate how this absolute constraint dictates the cardinality, channel location, and coherent pairing of $\mathcal{I}_{bias}$.

\subsection{Minimal Cardinality of the Bias Set $\mathcal{I}_{bias}$}

To reliably form an attention sink, the post-RoPE contribution $S_{post, bias}(\Delta m)$ must remain massively positive across a wide range of sequence distances $\Delta m \ge 1$. Given a fixed structural bias energy budget $E_{total} = \sum_{i \in \mathcal{I}_{bias}} \|\mathbf{q}_i\|_2 \|\mathbf{k}_i\|_2$, the network must optimize the energy allocation within $\mathcal{I}_{bias}$ to maximize this score. We demonstrate that avoiding positional variance mathematically forces $\mathcal{I}_{bias}$ to have minimal cardinality, ideally restricting it to a single element ($|\mathcal{I}_{bias}| = 1$).

Consider the difference between concentrating this energy versus distributing it. If the entire energy $E_{total}$ is assigned to a single subspace $i^*$, the contribution forms a cohesive wave:
\begin{footnotesize}
    \begin{equation}
    \label{eq:single_wave}
        S_{post, bias}(\Delta m) = E_{total} \cos(\alpha_{i^*} - \Delta m \cdot \theta_{i^*})
    \end{equation}
\end{footnotesize}
With only one rotational frequency present, the structural bias completely avoids internal conflict, fully utilizing the energy magnitude $E_{total}$.

Conversely, if the energy is distributed across two subspaces ($E_a + E_b = E_{total}$) with distinct frequencies $\theta_a \neq \theta_b$, the score becomes a superposition of mismatched waves:
\begin{footnotesize}
    \begin{equation}
        \begin{split}
            S_{post, bias}(\Delta m) = & E_a \cos(\alpha_a - \Delta m \cdot \theta_a) \\
            & + E_b \cos(\alpha_b - \Delta m \cdot \theta_b)
        \end{split}
    \end{equation}
\end{footnotesize}
Because RoPE frequencies strictly depend on the channel index, the two components evaluate at different rotational speeds. As sequence distance $\Delta m$ increases, the two waves inevitably drift out of phase. By the triangle inequality, the superposition of misaligned vectors is strictly sub-additive. Consequently, $|S_{post, bias}(\Delta m)| < E_{total}$ for distances where the phases mismatch, causing destructive interference that continuously suppresses the attention score.

Allocating the bias to more than two subspaces exacerbates this degradation. Superimposing additional distinct frequencies ($\theta_c, \theta_d, \dots$) accelerates phase mismatch and guarantees catastrophic phase cancellation at various sequence distances. This creates severe oscillations that destroy the reliable, massive baseline required for a global attractor.

To maximize magnitude and eliminate internal rotational variance, the model is forced to avoid frequency mixing entirely. This constraint explains the empirical observation that the size of $\mathcal{I}_{bias}$ is extremely small, with the model restricting the structural bias to \emph{a single active subspace} most of the time \citep{SystematicOutliers}.

\subsection{Low-Frequency Localization of the Bias Subspace}

Having established that $\mathcal{I}_{bias}$ optimally reduces to a single dominant subspace $i^*$, the post-RoPE bias contribution simplifies to a single wave as shown in Equation \ref{eq:single_wave}. To function as a reliable global attractor, this score must remain invariant across varying sequence distances $\Delta m \ge 1$. A large rotational frequency $\theta_{i^*}$ would induce sequence-dependent phase shifts, continuously degrading the structural bias as $\Delta m$ increases. To preserve the maximum magnitude $E_{total}$ regardless of the query's initial geometric alignment $\alpha_{i^*}$, the positional penalty must be strictly minimized:
$$\Delta m \cdot \theta_{i^*} \approx 0 \implies \theta_{i^*} \approx 0$$

In the standard RoPE formulation, rotational frequencies decay exponentially with respect to the channel index ($\theta_i = b^{-2i/d}$). Consequently, neutralizing this rotational penalty mathematically forces the network to localize the bias subspace $i^*$ within high-index channels. At these elevated indices, the frequency $\theta_{i^*}$ asymptotically approaches $b^{-1} \approx 0$. This effectively "freezes" the rotational transformation, isolating the structural bias from positional perturbation and guaranteeing its stability across a long context window of varying $\Delta m$. 

\subsection{Coherent Pairing and Variance Cancellation}

To maintain the stable, invariant score $S_{post, bias}(\Delta m)$ established in the subsection above, the mechanism must evaluate the inner product across the targeted subspace without information loss. We demonstrate that avoiding artificial magnitude decay mathematically necessitates the concurrent activation of both channels, providing the mathematical foundation for our empirical observation of coherent pairing.

Let the bias vectors in the dominant subspace $i^*$ be defined as $\mathbf{q}_{i^*} = (q_{i^*}, q_{i^*+d/2})^\top$ and $\mathbf{k}_{i^*} = (k_{i^*}, k_{i^*+d/2})^\top$, subject to the rotation angle $\phi = \Delta m \cdot \theta_{i^*}$. The full post-RoPE contribution algebraically expands to:

\begin{small}
\begin{equation}
    \begin{split}
        S_{post, bias}(\Delta m) = \underbrace{q_{i^*} (k_{i^*} \cos\phi - k_{i^*+d/2} \sin\phi)}_{\text{Axis } i^*} \\ + \underbrace{q_{i^*+d/2} (k_{i^*} \sin\phi + k_{i^*+d/2} \cos\phi)}_{\text{Axis } i^*+d/2}
    \end{split}
\end{equation}
\end{small}

\paragraph{Single-Channel Degradation.} If the model relies on a 1D projection—for instance, activating only the first channel such that $q_{i^*+d/2} \approx 0$—the rotation introduces a first-order $-q_{i^*} k_{i^*+d/2} \sin\phi$ penalty. As sequence distance $\Delta m$ increases, energy rotating out of the $i^*$ axis becomes mathematically invisible to the query. This unrecoverable energy loss causes the evaluated structural bias to artificially decay, severely compromising the stability of the attention sink.

\paragraph{Coherent 2D Activation.} Conversely, when both channels are concurrently active, the orthogonal axis actively captures this displaced energy via the $+q_{i^*+d/2} k_{i^*} \sin\phi$ term. By geometrically aligning the bias vectors ($q_{i^*} k_{i^*+d/2} \approx q_{i^*+d/2} k_{i^*}$), the model ensures these cross-terms perfectly cancel each other out. This mathematical cancellation entirely deletes the $\sin\phi$ rotational variance.

Therefore, to further prevent positional degradation and guarantee that the attention sink remains a structurally rigid, invariant global attractor across the entire context window, it is encouraged that the model coherently activates both channels within the chosen subspace.
\section{Implementation Details of \ours}
\label{app:insertquant_details}

In this section, we provide the granular algorithms and implementation specifications for the \ours framework, covering the progressive detection strategy, template matching logic, the precise operational schedule, and the spike-aware quantization calibration scheme.

\subsection{Sink Token Detection}
\label{sec:app_detection}

\paragraph{Thresholding Methodology.}
Following \citet{Prefixquant}, we identify sink tokens based on their maximum activation magnitudes relative to the \textbf{current sequence statistics}. A token $\mathbf{x}_i \in \mathbb{R}^d$ is flagged as a candidate sink token if its infinity norm significantly exceeds the median peak magnitude within the current input sample.

Formally, for an input sequence of length $T$, let the set of token magnitudes be $\mathcal{M} = \{ \|\mathbf{x}_t\|_\infty \}_{t=1}^T$. The detection threshold $\tau$ is defined as:
\begin{equation}
    \tau = \alpha \cdot \text{median}(\mathcal{M})
\end{equation}
where $\alpha$ is a scaling hyperparameter. Any token satisfying $\|\mathbf{x}_i\|_\infty > \tau$ is identified as a sink. 

Additionally, as observed in prior works \citep{AttentionSink, MassiveActivations}, the first token invariably functions as an attention sink in autoregressive language models. Therefore, we explicitly include the first token (position 0) in the sink set $\mathcal{S}$ by default.

\paragraph{Progressive Detection Strategy.}
Following the methodology of \citet{Prefixquant}, we standardly identify sink tokens at the input of the Layer 1 Down-Projection using a scaling factor of $\alpha_{1}$. However, our empirical analysis reveals that precursors to these massive spikes are often distinguishable as early as the input to the \textbf{Layer 0 Down-Projection}. To maximize dynamic range reduction, we employ a \textbf{Progressive Detection} strategy:
\begin{enumerate}
    \item \textbf{Stage 1 (Layer 0):} We perform an initial scan at the input of the Layer 0 Down-Projection using another threshold ($\alpha_0$). Due to the nascent state of the activation patterns, only the subset of sink tokens exhibiting early-onset spikes are detected. The remaining sink tokens, whose spikes have not yet fully formed, are simply not detected (false negatives) at this stage. The early detection allows us to clamp these tokens earlier to constrain the activation dynamic range. 
    \item \textbf{Stage 2 (Layer 1):} We perform the definitive scan at the input of the Layer 1 Down-Projection (the standard emergence point). This stage captures any remaining sink tokens that did not spark in Layer 0.
\end{enumerate}
The union of indices detected in Stage 1 and Stage 2 forms the final set of sink token indices $\mathcal{S}$, which is frozen and shared for all subsequent layers.

\subsection{Template Extraction and Matching}
\label{sec:app_matching}

Since distinct sink tokens generate unique structural biases, they cannot be represented by a single global template. Instead, we construct a dictionary of templates corresponding to the specific functional roles identified in the model.

\paragraph{Dictionary Construction.}
We define the template set $\mathcal{T} = \{ \mathbf{t}_1, \dots, \mathbf{t}_K \}$, where $K$ is a \textbf{model-specific hyperparameter} determined by the number of distinct sink behaviors observed during calibration (e.g., $K=3$ for Mistral-7B). During the offline calibration phase, we extract these templates directly from the calibration samples. For each layer $l$ in the persistence phase, we collect tThe mean Key and Value vectors extracted directly from the corresponding attention head outputs.

\paragraph{Template Assignment Strategy.}
It is possible that the number of detected sink tokens ($N_{sink}$) in a given sequence differs from the expected template count ($K$), often when observed for sequences that miss low-semantic tokens ($N_{sink} < K$) or have duplicated low-semantic tokens ($N_{sink} > K$). To address this, we employ a hybrid assignment strategy based on the exact match of sink counts:
\begin{enumerate}
    \item \textbf{Sequential Alignment ($N_{sink} = K$):} In standard sequences where the number of detected sink tokens matches the template count exactly, we assign templates based on their sequential order (i.e., the first detected sink maps to $\mathbf{t}_1$, the second to $\mathbf{t}_2$, etc.). We hypothesize that template types naturally follow a consistent sequential order due to the causal nature of the attention mechanism and the position-dependent rotation applied by RoPE.
    \item \textbf{MSE-Based Fallback ($N_{sink} \neq K$):} In cases where the counts do not match, we resolve the mapping ambiguity by computing the Mean Squared Error (MSE) between the detected token's state $\mathbf{x}_s$ and the candidate templates. The token is assigned to the template $k^*$ that minimizes the reconstruction error:
    \begin{equation}
        k^* = \arg\min_{k} \|\mathbf{x}_s - \mathbf{t}_k^{(detection)}\|_2^2
    \end{equation}
\end{enumerate}

\paragraph{Runtime Persistence.}
After the detection phase, the identified sink token indices $\mathcal{S}$ and their corresponding template assignments $\{k^*_s\}_{s \in \mathcal{S}}$ are shared across all subsequent layers $l > 1$. This ensures that the inserted templates consistently correspond to the assigned functional role of each token throughout the network depth. The above strategy applies not only to offline calibration but also runtime evaluation.

\subsection{Spike-aware Calibration for Quantization Parameters}
\label{sec:app_calibration_params}

Standard static quantization often utilizes grid search to find optimal clipping thresholds that minimize the Mean Squared Error (MSE) between quantized and full-precision activations \citep{Prefixquant}. However, applying this directly to \ours is suboptimal due to the deliberate removal of sink tokens. We utilize a \textbf{Structure-Constrained Search} to find the optimal clipping factors, introducing two critical adaptations:

\paragraph{1. Semantic-Only MSE Objective.}
Since \ours intentionally clamps sink tokens to zero—introducing a massive theoretical error at the current layer that is corrected via insertion in the subsequent layer—including these tokens in the loss calculation would skew the optimization. Instead, we compute the reconstruction error exclusively over the semantic tokens ($X_{semantic}$), effectively ignoring the temporary clamping artifacts on the sink tokens:
\begin{equation}
    \alpha^* = \arg\min_{\alpha} \text{MSE}(X_{semantic}, Q_\alpha(X_{semantic}))
\end{equation}

\paragraph{2. Structure Preservation Constraint.}
Aggressive clipping can suppress activation magnitudes, potentially altering the set of tokens identified as spikes by our detection algorithm (false negatives). To prevent this, we constrain the search space. We define the \textit{detection correctness} $R_{correct}$ as:
\begin{equation}
    R_{correct} = \frac{|\mathcal{S}_{quant} \cap \mathcal{S}_{fp}|}{|\mathcal{S}_{fp}|}
\end{equation}
where $\mathcal{S}_{quant}$ and $\mathcal{S}_{fp}$ are the sets of detected sink token indices in the quantized and full-precision calibration sets, respectively. We reject any clipping factors where $R_{correct}$ falls below a tolerance threshold of $0.8$, ensuring that the structural integrity of the sink mechanism is maintained even under low-bit quantization.
\section{Sensitivity Analysis of Prefix-Based Spike Suppression}
\label{app:prefixquant_robustness}

While prefix-based methods \citep{CushionCache, Prefixquant} effectively mitigate spikes in many standard settings, our stress-testing reveals specific boundary conditions where the heuristic assumption—that spikes only occur in the first instance of a token—becomes less robust. 

\begin{figure*}[ht]
    \centering
    \begin{minipage}{0.47\textwidth}
        \centering
        \includegraphics[width=\linewidth]{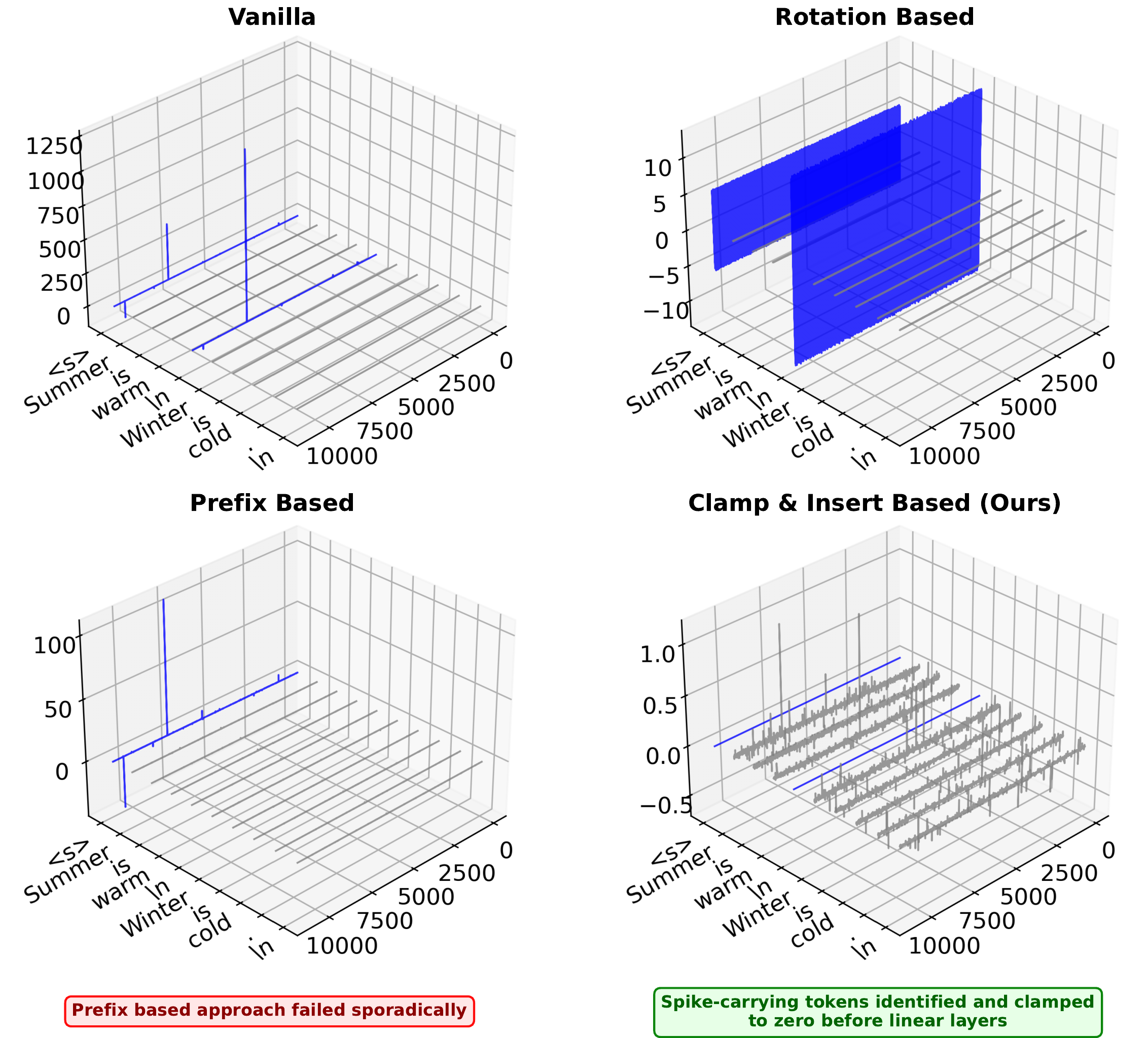}
        \caption{Rotation-based method amortizes spike across channels, but the token-wise extremes remain. Prefix-based method does not guarantee spike elimination when a \texttt{<BOS>} token exists in the start of sequence.}
        \label{fig:heatmap}
    \end{minipage}
    \hfill 
    \begin{minipage}{0.45\textwidth}
        \centering
        \includegraphics[width=\linewidth]{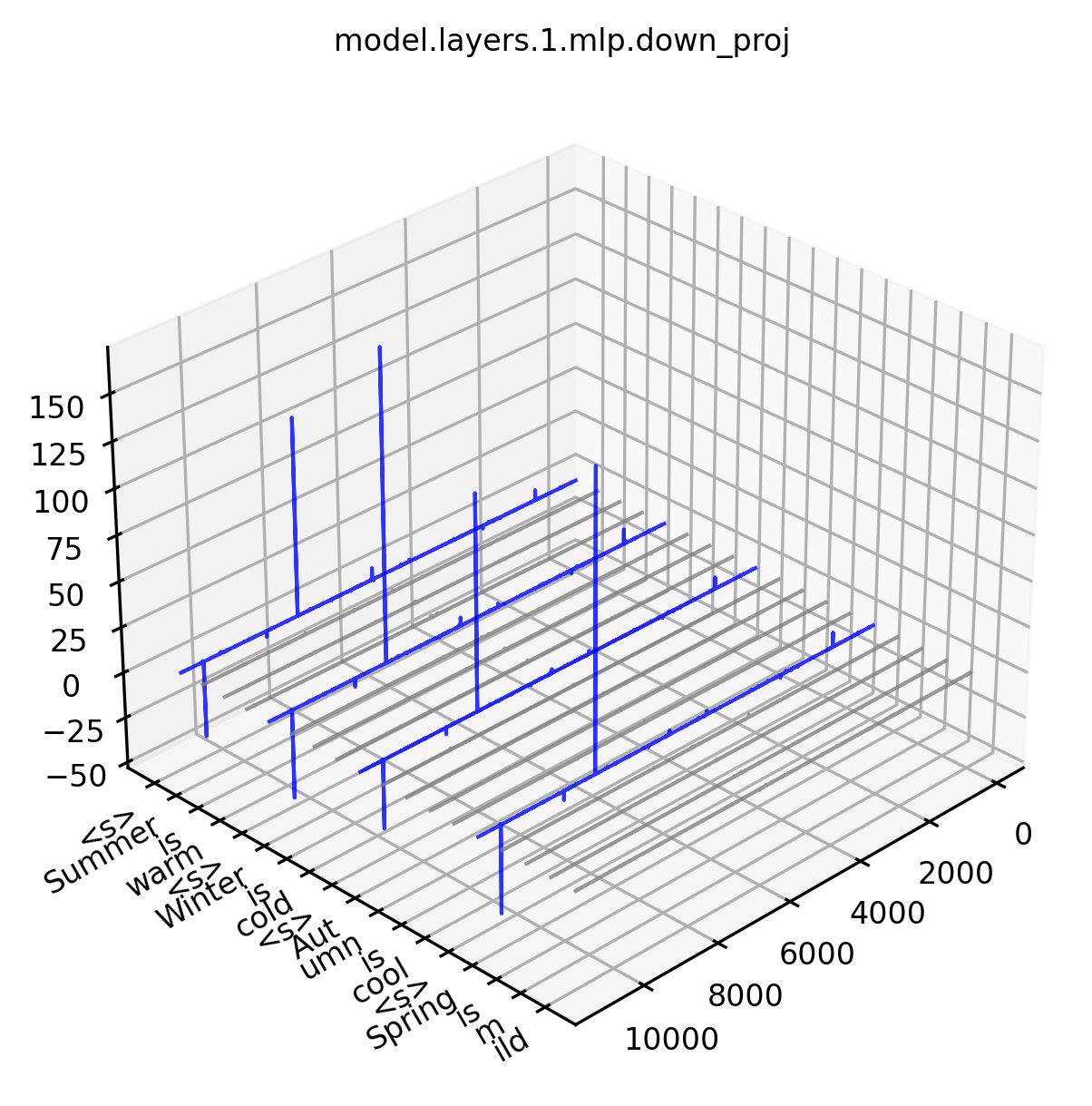}
        \caption{\texttt{<BOS>} can trigger spikes even when it is already prefixed in the KV cache. This might lead to suboptimal quantization quality for sequence packing, where inputs are concatenated to the same sequence.}
        \label{fig:bos}
    \end{minipage}
\end{figure*}

\paragraph{Residual Spikes in Recurrent Tokens}A core premise of prefix-based suppression is that once a "sink token" (e.g., \texttt{<BOS>}) is introduced at the start of a sequence, subsequent instances of that same token will function as normal, non-spiking tokens. However, Figure \ref{fig:heatmap} suggests that this suppression is not always absolute. When a \texttt{<BOS>} token exists in the start of the sequence with \texttt{<BOS>} already prefixed to KV cache, the second instance still retains a non-negligible spike magnitude. While reduced compared to the initial sink, these residual spikes can still exceed the quantization dynamic range (e.g., remaining $100\times$ larger than the average activation), potentially degrading precision in low-bit settings. This indicates that the "sink" role is not strictly binary but may be partially distributed across recurrent instances. While user can always skip the \texttt{<BOS>} if it strictly reside in the start of the sequence, it can lead to suboptimal quantization performance for when multiple \texttt{<BOS>} can be present (as shown in Figure \ref{fig:bos}), such as sequence packing.

\section{Detailed Experimental Configuration}
\label{app:exp_config}

\subsection{Quantization Configuration}
We utilize standard quantization schemes to ensure a fair comparison across all baselines. Unless otherwise specified, the configuration is as follows:

\begin{itemize}
    \item \textbf{Weights:} Symmetric per-channel quantization.
    \item \textbf{Activations:} Asymmetric quantization.
    \begin{itemize}
        \item \textit{Dynamic:} Per-token granularity. Activation clipping is not applied.
        \item \textit{Static:} Per-tensor granularity. Activation clipping is applied here, as this does not incur any runtime overhead.
    \end{itemize}
    \item \textbf{KV Cache:} Asymmetric quantization.
    \begin{itemize}
        \item \textit{Dynamic:} Group-wise granularity (group size 128).
        \item \textit{Static:} Per-head granularity.
    \end{itemize}
\end{itemize}

\subsection{MSE-Based Clipping Search}
For static activation quantization, we determine the optimal clipping factor via grid search to minimize layer-wise MSE using 8 calibration samples (sequence length 2048) from the PILE dataset using the method from \citet{Prefixquant}. For \ours, we implement two necessary adaptations:

First, the clamped outlier tokens inherently exhibit high error at the current layer's output (an intentional artifact that is corrected via template insertion in the subsequent layer). we compute reconstruction error exclusively over semantic tokens, ignoring the temporary clamping error of spikes. Second, overly aggressive clipping can suppress activation magnitudes, causing the spike detection to miss valid spike-carrying tokens. we constrain the search space to reject any clipping factors that lead to wrong spike-carrying token positions detection in the calibration set, ensuring that the structural integrity is not compromised.
\section{Documentation of Assets and Licenses}
\label{appendix:assets}

We provide documentation for all external models, datasets, and software frameworks utilized in this work.

\subsection{Pretrained Models}
All models were obtained via the Hugging Face Model Hub or official repositories. We utilized specific versions of these models as baselines for our quantization and mechanistic simulations.

\begin{table}[h]
\centering
\resizebox{\columnwidth}{!}{%
\begin{tabular}{@{}lll@{}}
\toprule
\textbf{Model} & \textbf{License} & \textbf{Source} \\ \midrule
LLaMA-2-7B \cite{llama2} & Llama 2 Community & \href{https://huggingface.co/meta-llama/Llama-2-7b-hf}{[HuggingFace]} \\
LLaMA-2-70B \cite{llama2} & Llama 2 Community & \href{https://huggingface.co/meta-llama/Llama-2-70b-hf}{[HuggingFace]} \\
LLaMA-3.2-3B \cite{llama3.2} & Llama 3.2 Community & \href{https://huggingface.co/meta-llama/Llama-3.2-3B}{[HuggingFace]} \\
Mistral-7B-v0.3 \cite{mistral7bv03} & Apache 2.0 & \href{https://huggingface.co/mistralai/Mistral-7B-v0.3}{[HuggingFace]} \\
CLIP ViT-L \cite{CLIP} & MIT & \href{https://huggingface.co/openai/clip-vit-large-patch14}{[HuggingFace]} \\
DINOv2 ViT-L \cite{dinov2} & Apache 2.0 & \href{https://huggingface.co/facebook/dinov2-large}{[HuggingFace]} \\ DINOv2 ViT-B \cite{dinov2} & Apache 2.0 & \href{https://huggingface.co/facebook/dinov2-base}{[HuggingFace]} \\ \bottomrule
\end{tabular}%
}
\caption{Documentation of pretrained language and vision models used in this study.}
\end{table}

\subsection{Datasets}
We evaluate our methods across the following benchmarks and calibration datasets:
\begin{itemize}
    \item \textbf{Calibration}: The Pile \cite{pile} (MIT).
    \item \textbf{Language}: WikiText-2 \cite{wikitext} (CC BY-SA 3.0).
    \item \textbf{Reasoning}: HellaSwag \cite{HellaSwag} (MIT), ARC (Easy/Challenge) \cite{ARC} (CC BY-SA 4.0), BoolQ \cite{BoolQ} (CC BY-SA 3.0), PIQA \cite{PIQA} (Apache 2.0), OpenBookQA \cite{openbookQA} (Apache 2.0), Winogrande \cite{winogrande} (Apache 2.0), and SocialIQA \cite{SocialIQA} (CC BY 4.0).
    \item \textbf{Vision}: ImageNet-1K \cite{ImageNet-1K} (Custom; Non-Commercial), and Flickr30k \cite{flickr30k} (Custom; Research Use) via \href{https://huggingface.co/datasets/nlphuji/flickr30k}{[HuggingFace]}.
\end{itemize}

\subsection{Software and Frameworks}
The following open-source frameworks were used for evaluation, model management, and system overhead measurement:
\begin{itemize}
    \item \textbf{lm-evaluation-harness}: Zero-shot evaluation. MIT License. \href{https://github.com/EleutherAI/lm-evaluation-harness}{[GitHub]}
    \item \textbf{HF Transformers}: Model management and emulation. Apache 2.0. \href{https://github.com/huggingface/transformers}{[GitHub]} \cite{HuggingFace}
    \item \textbf{Triton}: Custom kernel implementation for latency benchmarking. MIT License. \href{https://github.com/triton-lang/triton}{[GitHub]}
\end{itemize}

\subsection{License Compliance}
The authors certify that all assets were used in strict compliance with their respective licenses. We do not redistribute weights of models with restrictive community licenses.
\section{Model Size and Compute Resources}
\label{appx:compute-resources}

\paragraph{Model Sizes.} In this work, we evaluate our proposed \textsc{InsertQuant} framework across a diverse range of model architectures and scales. The evaluated language models include LLaMA-3.2 (3B parameters), LLaMA-2 (7B and 70B parameters), and Mistral-v0.3 (7B parameters). For vision architectures, we evaluate CLIP ViT-L (approx. 300M parameters) and DINOv2 (ViT-B at 86M parameters, and ViT-L at 300M parameters).

\paragraph{Compute Resources.} A primary advantage of our \textsc{InsertQuant} PTQ framework is its minimal computational overhead. Because our method relies on a lightweight offline calibration phase (using only 64 sequences from the Pile) and surgical intervention via template insertion, we do not require the massive distributed clusters typically necessary for Quantization-Aware Training (QAT).

All offline template calibration, spike detection, and downstream zero-shot evaluations for 3B and 7B models can be fully executed on a single compute node equipped with one NVIDIA A100 GPU. For the larger LLaMA-2-70B model, due to its substantial memory requirements, these processes were executed across four NVIDIA A100 GPUs. While we distributed some parallel evaluation sweeps for the smaller models across multiple GPUs to accelerate experimental turnaround times, a single GPU is strictly sufficient for their reproduction.

\paragraph{Computational Budget.} Evaluating the quantized models on the downstream common-sense reasoning, perplexity, and vision benchmarks required approximately 1 to 2 GPU hours per model configuration. We estimate the total compute required to generate the final accuracy and perplexity tables reported in this manuscript to be roughly 30 GPU hours. Factoring in preliminary experiments, latency benchmarking, failed configurations, and exploratory parameter sweeps during the development phase, the total compute footprint for this research project was approximately 100 GPU hours.

\end{document}